\documentclass[10.5pt,twocolumn,journal,letterpaper]{IEEEtran}
\usepackage{amsmath, amsthm, amssymb}
\usepackage{url,flushend,multirow,booktabs}
\usepackage{algorithm,algorithmic}   % no crossline
\usepackage{graphicx}
\usepackage{bm}
\usepackage{cite}
\usepackage{subfigure}
\usepackage{float}
\usepackage{color}
\usepackage{multirow}

\usepackage{colortbl}
\definecolor{maroon}{cmyk}{0.08,0.04,0.00,0.06}  % light blue
\definecolor{citecol}{RGB}{0,220,0}  % green, RGB [0-255], rgb: 0-1

\usepackage[colorlinks,citecolor=citecol,urlcolor=blue,bookmarks=false,hypertexnames=true]{hyperref} % ,citecolor=green
\newcommand{\eg}{e.g.}
\newcommand{\ie}{i.e.}

\newcommand{\etal}{\textit{et al.}}

\renewcommand{\bm}{\mathbf}
\DeclareFixedFont{\mf}{OT1}{ptm}{m}{n}{10pt}
\DeclareFixedFont{\mfb}{OT1}{ptm}{bx}{n}{10pt}

\begin{document}
%

\title{Pair-wise Layer Attention with Spatial Masking for Video Prediction}

\author{Ping~Li, Chenhan~Zhang, Zheng~Yang, Xianghua~Xu, and Mingli~Song,~\IEEEmembership{Senior Member,~IEEE}
%\thanks{Manuscript received November 16th, 2023.}
\thanks{P.~Li, C.~Zhang and X.~Xu are with the School of Computer Science and Technology, Hangzhou Dianzi University, Hangzhou, China (e-mail: \{lpcs, zch2020, xhxu\}@hdu.edu.cn). P.~Li is also with Guangdong Laboratory of Artificial Intelligence and Digital Economy (SZ), China.}
%\thanks{L.~Yuan is with the School of Electronic and Computer Engineering, Peking University, China, and also with PengCheng Laboratory, China (e-mail:yuanli-ece@pku.edu.cn).}
\thanks{Z.~Yang is with FABU Technology Co., Ltd., Hangzhou, China (e-mail:yangzheng@fabu.ai)}
\thanks{M.~Song is with the College of Computer Science, Zhejiang University, Hangzhou, China. (e-mail:songml@zju.edu.cn).} 
}
\markboth{Draft}
{LI \MakeLowercase{\textit{et al.}}:~Pair-wise Layer Attention with Spatial Masking for Video Prediction}

\maketitle

\begin{abstract}
  Video prediction yields future frames by employing the historical frames and has exhibited its great potential in many applications, \eg, meteorological prediction, and autonomous driving. Previous works often decode the ultimate high-level semantic features to future frames without texture details, which deteriorates the prediction quality. Motivated by this, we develop a Pair-wise Layer Attention (PLA) module to enhance the layer-wise semantic dependency of the feature maps derived from the U-shape structure in Translator, by coupling low-level visual cues and high-level features. Hence, the texture details of predicted frames are enriched. Moreover, most existing methods capture the spatiotemporal dynamics by Translator, but fail to sufficiently utilize the spatial features of Encoder. This inspires us to design a Spatial Masking (SM) module to mask partial encoding features during pretraining, which adds the visibility of remaining feature pixels by Decoder. To this end, we present a Pair-wise Layer Attention with Spatial Masking (PLA-SM) framework for video prediction to capture the spatiotemporal dynamics, which reflect the motion trend. Extensive experiments and rigorous ablation studies on five benchmarks demonstrate the advantages of the proposed approach. The code is available at \href{https://github.com/mlvccn/PLA_SM_VideoPred}{GitHub}. 
 
\end{abstract}


 \ifCLASSOPTIONpeerreview
 \begin{center} \bfseries EDICS Category: 3-BBND \end{center}
 \fi

\IEEEpeerreviewmaketitle

\section{Introduction}
\label{sec1:intro}

\IEEEPARstart{V}{ideo} prediction \cite{li-tmm2023-ffinet} employs the historical frames to generate the future frames, and it is regarded as a pixel-level unsupervised learning task. It has gained much popularity in widespread applications, \eg, autonomous driving, robot, and weather forecast. Usually, video prediction methods employ Recurrent Neural Network (RNN) \cite{yu-nc2019-rnn} as the backbone which captures the temporal dependency. For example, PredRNN (Predictive RNN) \cite{wang-nips2017-predrnn}, PredRNN++ \cite{wang-icml2018-predrnn++}, MIM (Memory In Memory) \cite{wang-cvpr2019-mim}, and MotionRNN (Motion Recurrent Neural Network) \cite{wu-cvpr2021-motionrnn} stack multiple RNNs to build the prediction model. 

To capture the spatial features, recent methods adopt the end-to-end two-stage feature learning framework, such as E3D-LSTM (Eidetic 3D LSTM) \cite{wang-iclr2019-e3dlstm}, CrevNet (Conditionally Reversible Network) \cite{yu-iclr2020-crevnet}, MAU (Motion-Aware Unit) \cite{chang-nips2021-mau}, and SimVP (Simple Video Prediction) \cite{gao-cvpr2022-simvp}. They use Convolutional Neural Network (CNN) \cite{li-tnnls2022-cnn} to extract spatial features and RNN to learn temporal features. These models consist of three main parts including Encoder (CNN), Translator (RNN), and Decoder (CNN). However, they focus on designing Translator, and use simple convolutions to build Encoder or Decoder when handling each frame individually. This leads to much redundancy and less diversity in spatial features, because those frames within the same video have similar background. 

Generally, typical video understanding tasks such as action recognition and temporal action detection \cite{li-pr2023-tad} predicts the results according to the features in the last layer, which is far from sufficient for the pixel-level video prediction task since the texture details become less and less when the network depth gets large, such as object contours and tiny objects. Although skip connection is used between Encoder and Decoder to provide low-level textures in previous works \cite{chang-nips2021-mau, chang-cvpr2022-strpm, gao-cvpr2022-simvp, guen-cvpr2020-phydnet}, the texture features are simply concatenated without the spatiotemporal update in Translator, resulting in misleading details of predicted frames. To overcome this drawback, we adopt the U-shape structure to stack the spatiotemporal update modules to build the symmetry Translator, \ie, the features of the $i$-th layer and that of the $(2N+1-i)$-th layer comply to the similar data distribution for a $2N$-layer Translator. In particular, we develop a Pair-wise Layer Attention (\textbf{PLA}) block to capture the layer-wise dependency using attention mechanism, which allows to capture the global context and incorporate high-level semantic features with low-level visual cues to facilitate future frame prediction. 

Inspired by masked autoencoders \cite{he-cvpr2022-mae} in self-supervised learning, we adopt the masking strategy in video prediction by randomly masking the pixel area of input frames during pretraining. This indeed helps to increase the feature diversity. However, it is computationally intensive because the GPU computations desired by the self-attention in mask autoencoders increase squarely with video length. To reduce computational costs, one can use fully convolutional neural networks due to its efficiency as indicated by \cite{gao-cvpr2022-simvp}, but the common convolutional kernel is sensitive to edges, which leads to distortions in the feature of masked image \cite{jing-arxiv2022-mscn}. To address this issue, we adopt sparse convolution \cite{liu-cvpr2015-sparseconv} to extract the features of masked image since it only computes the visible part of image, which alleviates the distortion problem to some degree. Hence, we design a Spatial Masking (\textbf{SM}) module that employs the attention mechanism to adaptively mask the spatial features derived from Encoder during pretraining, and this module is not used in training.

Therefore, we propose a Pair-wise Layer Attention with Spatial Masking framework for video prediction. Particularly, the spatial masking strategy is only used during pretraining to make the learned features more robust, while the pair-wise layer attention module captures high-level semantics and low-level details to enrich the global context among frames. To investigate the performance of our method, extensive experiments were conducted on several benchmarks including Moving MNIST \cite{srivastava-icml2015-movingmnist}, TaxiBJ \cite{zhang-aaai2017-trafficbj}, Human3.6M \cite{ionescu--tpami2014-human3.6m}, KITTI\&Caltech Pedestrian \cite{geiger-ijrr2013-kitti}, and KTH \cite{schuldt-icpr2004-kth}, whose results have well verified the advantage of the proposed approach.
 
The main contributions are summarized as follows:
\begin{itemize}
  \item A Pair-wise Layer Attention with Spatial Masking (\textbf{PLA-SM}) framework is developed for video prediction, which consists of three primary components, \ie, Encoder, PLA-based Translator, and Decoder.
  
  \item A Spatial Masking strategy is proposed to increase the robustness of encoding features, by randomly masking the frame features during pretraining. 
  
  \item The pair-wise layer attention mechanism is designed for capturing the spatiotemporal dynamics that reflects the motion trend in video, by simultaneously considering both high-level semantics and low-level detailed cues of frames.

\end{itemize}

%The rest of this paper is organized as follows. Section~\ref{related} reviews some closely related works and Section~\ref{method} introduces our video prediction framework. Then, we report the experimental results on several benchmarks to demonstrate the advantage of our method in Section~\ref{test}. Finally, we conclude this work in Section~\ref{conclusion}.
%

% --------------  PLA-SM Framework --------------
\begin{figure*}[!t]
	\centering
	\includegraphics[width=0.9\linewidth]{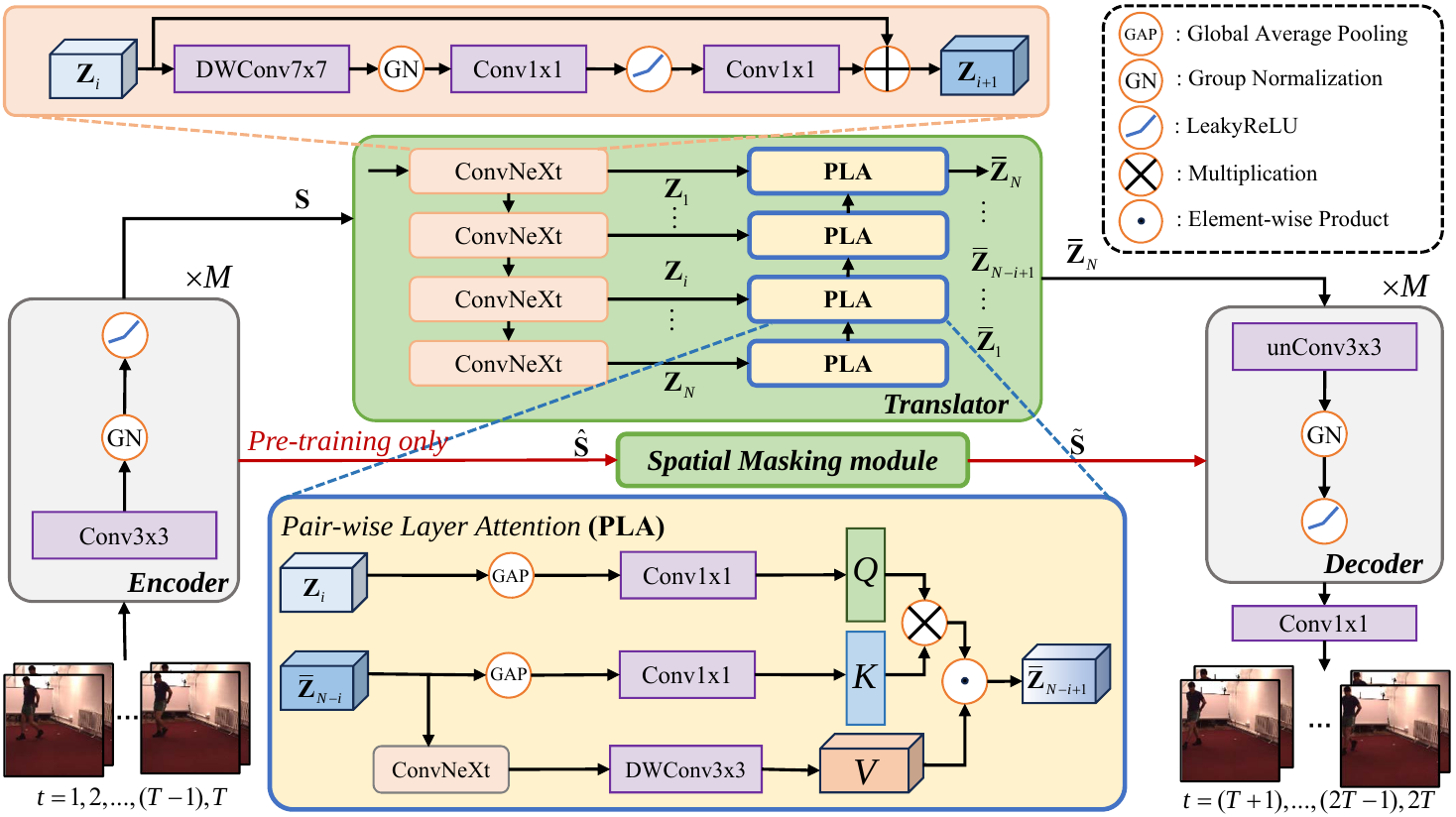}
	\caption{Overall framework of our PLA-SM method.}
	\label{fig:framework}
\end{figure*}

%-------------------------------------------------------------------------
\section{Related Work}
\label{related}
This section discusses the closely related works including video prediction and masked autoencoder. 

\subsection{Video Prediction} 
Early works \cite{ranzato-arxiv2014-videomodeling} adopt Recurrent Neural Network (RNN) to model the temporal dynamics in video. For example, Srivastava \etal~\cite{srivastava-icml2015-movingmnist} use the Long Short-Term Memory (LSTM) network with fully-connected (fc) layers, which require a large volume of parameters to learn; Shi \etal~\cite{shi-nips2015-convlstm} substitute fc layers with convolutional layers to reduce the model parameters. However, these methods update the memory only along the temporal dimension, which neglects the spatial dimension. To handle this drawback, Wang \etal~\cite{wang-nips2017-predrnn} proposed Predictive RNN (PredRNN) that uses the spatiotemporal flow to update the memory along both spatial and temporal dimensions; to model short-term dynamics by increasing the depth of recurrent units, PredRNN++ \cite{wang-icml2018-predrnn++} uses the causal LSTM with a cascaded mechanism and gradient highway unit to reduce the difficulty of gradient propagation; PredRNNv2 \cite{wang-tpami2023-predrnnv2} employs the reverse scheduled sampling method with the decoupling loss to enhance the context feature. Besides, Wang \etal~\cite{wang-cvpr2019-mim} developed the Memory In Memory (MIM) method to turn the time-variant polynomials into constant, thus making the deterministic component predictable; Oliu \etal~\cite{oliu-eccv2018-frnn} put forward a folded RNN to share state cells between Encoder and Decoder, thus reducing the costs. Unlike them, Motion RNN \cite{wu-cvpr2021-motionrnn} divides the physical motion into two parts, \ie, transient variation and motion trend, where the latter is regarded as the accumulation of previous motions. Unluckily, the above RNN methods fail to directly employ the frame features in the past but indirectly borrow the memory, which deteriorates the performance of the long-term video prediction. Hence, Su \etal~\cite{su-nips2020-convttlstm} update the features at the current time by using that of the previous continuous time instances, and generalize convolutional LSTM from the first-order Markov chain model to the higher-order one. All these methods build the video prediction model by stacking multiple RNNs while coupling the spatial and the temporal feature learning. 

Recent methods usually adopt the two-stage strategy, \ie, learn spatial features and temporal features by CNN and RNN respectively, which are implemented in an end-to-end way, such as Eidetic 3D LSTM (E3D-LSTM) \cite{wang-iclr2019-e3dlstm} and Conditionally Reversible Network (CrevNet) \cite{yu-iclr2020-crevnet}. The former uses 3D convolution to extract spatiotemporal features and combines it with RNN to learn motion-aware short-term features; the latter employs the reversible architecture to build a bijective two-way autoencoder and its complementary recurrent predictor. They are able to learn better spatial features, but it remains insufficient for accurately predicting the motion trend in complex scenes. To overcome this shortcoming, the Physical Dynamics Network (PhyDNet) \cite{guen-cvpr2020-phydnet} and the Video generation with Ordinary Differential Equation method (Vid-ODE) \cite{park-aaai2021-vidode} model the physical dynamics of objects from the partial and the ordinary differential equation perspectives; the Dynamic Motion Estimation and Evolution (DDME) \cite{kim-acmmm2021-dmee} generates different convolution kernels for video frames at each time instance to capture temporal dynamics. Moreover, some works focus on improving the performance of long-term video prediction, \eg, LMC (Long-term Motion Context) \cite{lee-cvpr2021-lmc} adopts the LMC memory with the alignment scheme to store the long-term context of training data and model the motion context; MAU (Motion-Aware Unit) \cite{chang-nips2021-mau} employs the history temporal features to enlarge the temporal receptive field; SimVP (Simple Video Prediction) simply uses the common convolutional modules to build a simple prediction model that achieves the SOTA performance. 

In addition, Chang \etal~\cite{chang-cvpr2022-strpm} proposed the spatiotemporal residual predictive model for high-resolution video prediction, which employs three encoder-decoder pairs to extract features with more details; Chen \etal~\cite{chen-cvpr2022-cpl} explored the continual learning in video prediction, and developed the mixture world model with the predictive experience-replay strategy to alleviate the catastrophic forgetting problem. Furthermore, Yu \etal~\cite{yu-cvpr2022-mac} introduced the semantic action-conditional video prediction, which is considered as the inverse action recognition, \ie, predict the video when the semantic labels that describes the interactions are available. 

Most of the above works use the common convolutions to build Encoder and Decoder without additional pretraining, but the designs of both Encoder and Decoder are far from achieving satisfying prediction performance.

\subsection{Masked Autoencoder} 
In natural language processing, BERT \cite{devlin-naccl2019-bert} removes some words of the document during training and recovers those removed words by the model. Such process is called Masked Auto-Encoding (MAE), which makes the model be equipped with stronger ability of capturing the context, and is successful in the computer vision field. For example, He \etal~\cite{he-cvpr2022-mae} masked a large portion of the image during pretraining and recovered the masked image by the model; Wei \etal~\cite{wei-cvpr2022-maskfit} employed the Histogram of Oriented Gradient (HOG) instead of image as the reconstruction goal of pretraining, which brings about better generalization ability. For the popular Vision Transformer (ViT), Gao \etal~\cite{gao-nips2022-mcmae} proposed the multi-scale hybrid convolution-transformer to learn more discriminative representations using MAE, and employed the masked convolution to prevent information leakage. For self-supervised representation pretraining, Chen \etal~\cite{chen-ijcv2023-cae} presented the Context AutoEncoder (CAE) to model masked images, including masked representation prediction and masked patch reconstruction. Besides, Tong \etal~\cite{tong-nips2022-videomae} applied MAE to videos with a higher masking rate. The mask modeling performs well on these ViT-based methods, but fails on CNN \cite{jing-arxiv2022-mscn} since masking may lead to the collapse of convolution kernels. To address this issue, Woo \etal~\cite{woo-cvpr2023-convnextv2} treated masked image as sparse data and employed sparse convolution to learn spatial features. While MAE has exhibited its great potential in many vision tasks, it still remains untouched in video prediction.

%-------------------------------------------------------------------------
\section{The Proposed Method}
\label{method}
This section describes the proposed Pair-wise Layer Attention with Spatial Masking (PLA-SM) framework, and we begin with the problem definition.

% --------------  Spatil Masking Framework --------------
\begin{figure}[!t]
	\centering
	\includegraphics[width=\linewidth]{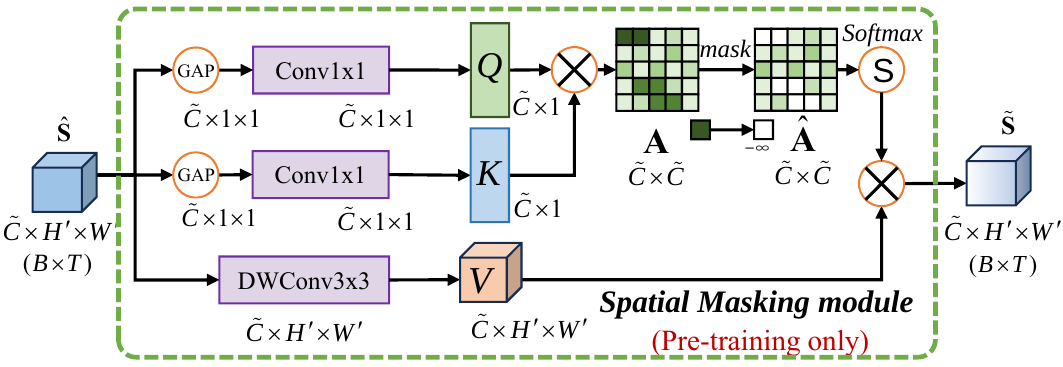}
	\caption{Architecture of Spatial Masking (SM) module.}
	\label{fig:sm_module}
\end{figure}

\subsection{Problem Definition}
Video prediction attempts to yield the future frames according to the historical frames. Given a video sequence with $T$ frames, \ie, $\mathcal{X}_T=\{\mathbf{X}_t\}^T_{t=1}$, its goal is to output the coming video sequence, \ie,  $\mathcal{Y}_{T^{\prime}}=\{\mathbf{X}_t\}^{T+T^\prime}_{t=T+1}$, where $\mathbf{X}_t\in\mathbb{R}^{C\times H \times W}$ is the $t$-th frame with channel $C$, height $H$, and width $W$. Mathematically, video prediction model learns a mapping function with learnable parameter $\Theta$, \ie, $\mathcal{F}_\Theta:\mathcal{X}_T \mapsto \mathcal{Y}_{T^\prime}$ and minimizes the loss function $\mathcal{L}(\cdot)$, \ie, 
\begin{equation}
	\Theta^* = \arg \min\limits_{\Theta}\mathcal{L}(\mathcal{F}_\Theta(\mathcal{X}_T),\mathcal{Y}_T^{\prime}),
\end{equation}
where the loss function adopts Mean Absolute Error (MAE) or Mean Square Error (MSE), and $\mathcal{F}_\Theta(\mathcal{X}_T)$ denotes the predicted frame.

\subsection{Overall Framework}
As illustrated in Fig.~\ref{fig:framework}, our video prediction framework consists of Encoder, Translator (PLA), and Decoder. During pretraining, Translator is substituted by the Spatial Masking module to enhance the features derived from Encoder as shown in Fig.~\ref{fig:sm_module}. 

Among them, Encoder is composed of stacked convolutions and group normalization with leaky ReLU as the activation function, which are used to learn spatial features of frames. Note that we use the sparse convolution \cite{liu-cvpr2015-sparseconv} to facilitate mask pretraining. Translator mainly consists of ConvNeXt \cite{liu-cvpr2022-convnext} and PLA, where the former captures the temporal dynamics in video, while the latter employs the low-level features with more textures to enhance the high-level features, and updates the spatiotemporal features to model the spatiotemporal dynamics that indicate the motion trend in future. Decoder includes stacked transposed convolutions and group normalization with leaky ReLU to decode the spatiotemporal features into predicted frames. The details are elaborated below.

\subsection{Encoder}
For Encoder, we stack $M$ convolution blocks $Conv(\cdot)$, each of which contains $3\times 3$ convolution and group normalization, followed by leaky ReLU that adds feature nonlinearity. Here, convolution layer learns the spatial feature of frames, which are then normalized to unit length along the feature channel to speedup the convergence of model. Note that the stride is set to 2 at every two convolution layers (the rest are 1), which actually does the feature downsampling, \ie, $H \rightarrow \frac{H}{2}$, $W \rightarrow\frac{W}{2}$.

During pretraining, it requires masking a large proportion of frame and common convolutions degrade the learning performance \cite{tian-iclr2023-spark} when there exist large masks. Hence, we adopt the sparse convolution to substitute the common one, which makes the model only compute the non-masking pixels and will not affect those masking ones. This helps to learn the spatial features from the non-masking area by employing a hashing table to record the positions of non-zero entries, and these features are then are recovered to frames. 

Formally, the input of Encoder is $B$ batches of video frame sequence $\{\mathcal{X}_T^{1}, \cdots, \mathcal{X}_T^{B}\}$, which are reshaped to the tensor $\tilde{\mathcal{X}_T}\in \mathbb{R}^{(B\cdot T)\times C\times H\times W}$, such that Encoder focuses on the spatial feature learning while neglecting the temporal dynamics. Then, these spatial features are concatenated along the temporal dimension, resulting in the spatial representation $\mathbf{S}\in \mathbb{R}^{B \times (T \cdot \tilde{C}) \times H^{\prime} \times W^{\prime}}$, where $\tilde{C}=64$, $H^\prime = H/2^{\lfloor \widetilde{N}/2 \rfloor}, W^\prime = W/2^{\lfloor \widetilde{N}/2 \rfloor}$ are channel number, height, and width respectively. 

\subsection{Translator}
Translator is composed of $N$ ConvNeXt blocks and PLA blocks, which are used to learn and update the spatiotemporal representation of video. It accepts the spatial representation $\mathbf{S}$ derived from Encoder as the input. 

\textbf{ConvNeXt}. It contains $7\times 7$ depth-wise convolution $DWConv(\cdot)$ \cite{howard-arxiv2017-mobilenet}, two $1\times 1$ 2D convolutions $Conv(\cdot)$, group normalization $GN(\cdot)$ and the activation function leaky ReLU $LReLU(\cdot)$ with a skip connection. Denoting the input of the $i$-th ($1\le i\le N$) ConNeXt block by $\bm{Z}_{i}\in \mathbb{R}^{B \times \hat{C} \times H^\prime \times W^\prime}$, where $\hat{C}=512$, it outputs the spatiotemporal representation $\mathbf{Z}_{i-1}$ with the same size. Note that the initial input is $\bm{Z}_0 = \mathbf{S}$. Mathematically, the computing process is expressed by
\begin{equation}
	\mathbf{Z}_i = \mathbf{Z}_{i-1} + Conv(LReLU(Conv(GN(DWConv(\bm{Z}_{i-1}))))),	
\end{equation}
where $DWConv(\cdot)$ \cite{howard-arxiv2017-mobilenet} accepts the convolution kernel with a larger size as it reduces the convolution parameters by decoupling the spatial dimension and the channel dimension. Specifically, it uses one convolution kernel for each channel and then concatenates the outputs using all convolution kernels. It enlarges the spatial receptive field by using larger kernel size compared to the common convolution with the same parameters. Here, $GN(\cdot)$ is used to speedup the convergence of model during training; $LReLU(\cdot)$ adds the nonlinearity of the learned representation; two $Conv(\cdot)$ layers constitute the inverted bottleneck structure, which changes the channel number twice, \ie, $\hat{C}\rightarrow 4\hat{C} \rightarrow \hat{C}$. The first convolution layer enriches the learned features and the second one reduces the redundancy by reducing the channel dimension.

\textbf{Pair-wise Layer Attention}. 
As depicted in Fig.~\ref{fig:framework}, Translator is designed to include more low-level texture cues in high-level spatiotemporal features by adopting U-Net like architecture, \ie, the $i$-th ConvNeXt block $ConvNeXt(\cdot)$ corresponds to the $(N-i+1)$-th PLA block $PLA(\cdot)$. Here the top-down ConvNeXt blocks are regarded as low-level layers while the bottom-up PLA blocks are treated as high-level layers. The basic idea is to employ the low-level cues, \ie, the output of ConvNeXt layer, to compensate for the corresponding high-level features, \ie, the output of PLA layer. That is why we name the block from the pair-wise layer perspective.

Each PLA block has two inputs, \ie, the output $\bm{Z}_i$ of the $i$-th ConvNeXt block (by skip connection) and the output $\bar{\bm{Z}}_{N-i}$ of the $(N-i)$-th PLA block, except for the first PLA block when $\bar{\bm{Z}}_0 = \bm{Z}_N$. The final output $\bar{\bm{Z}}_N$ is the input of Decoder. Here, we adopt the attention mechanism in terms of pair-wise layer. In particular, we do Global Average Pooling (GAP) operation on the ConvNeXt output $\bm{Z}_{i}\in \mathbb{R}^{B \times \hat{C} \times H^\prime \times W^\prime}$ (spatiotemporal feature) to obtain the global feature, which is then reduced to \textit{Query} feature $\bm{Q}_{i}\in \mathbb{R}^{B \times \hat{C} \times 1}$ by $1\times 1$ convolution and the squeeze operation (\ie, $\hat{C}\times 1\times 1 \rightarrow \hat{C}\times 1$). Similarly, we obtain the \textit{Key} feature $\bm{K}_{N-i}\in \mathbb{R}^{B \times \hat{C} \times 1}$ by using the same operations on the PLA output $\bar{\bm{Z}}_{N-i}\in \mathbb{R}^{B \times \hat{C} \times H^\prime \times W^\prime}$ (\textit{enhanced} spatiotemporal feature). This module aims to build the semantic relations between \textit{Key} feature and \textit{Query} feature, \ie, seek the similar area of the PLA high-level features to that of ConvNeXt low-level features. 

Meanwhile, the PLA output $\bar{\bm{Z}}_{N-i}$ is fed into one ConvNeXt block and one $3\times 3$ depth-wise convolution $DWConv(\cdot)$, resulting in the \textit{Value} feature $\bm{V}_{N-i}\in \mathbb{R}^{B \times \hat{C} \times H^\prime \times W^\prime}$, \ie, 
\begin{equation}
	\label{eq:value}
	\bm{V}_{N-i} = DWConv(ConvNeXt(\bar{\bm{Z}}_{N-i})),
\end{equation}  
where $DWConv(\cdot)$ is used to enrich the spatiotemporal features from the ConvNeXt block. To evaluate the channel importance, we compute the attention score $\bm{A}$ of low-level texture feature and high-level enhanced feature. When computing the attention score $\bm{A}_{(i,N-i)}$ of \textit{Query} and \textit{Key}, we omit the batch size $B$, \ie, $\bm{A}_{(i,N-i)} = Softmax(\bm{Q}_i^\top \bm{K}_{N-i}/\sqrt{\hat{C}})$, where $Softmax(\cdot)$ is a normalization function which makes the dot production value fall in between 0 and 1, and the denominator helps to avoid too large value. Note that we have attempted to use three ways to compute the attention score $\bm{A}$, including the channel attention ($[H\cdot W]\times [H\cdot W]$), the spatial attention ($\hat{C}\times \hat{C}$), and the global attention ($1\times 1$) schemes. Among them, the global attention strategy was empirically found to perform the best using the least cost. During the training, the model is expected to adjust the weight of feature according to the channel importance score. 

Having obtained \textit{Query}, \textit{Key}, and \textit{Value}, we form an attention tensor with the channel importance score to the same size of \textit{Value} by repeating the score, and compute the element-wise product between them, resulting in the enhanced spatiotemporal feature $\bar{\bm{Z}}_{N-i+1}\in \mathbb{R}^{B \times \hat{C} \times H^\prime \times W^\prime}$. In this way, the high-level features are enhanced layer by layer, and the output of the top PLA block is the spatiotemporal feature $\bar{\bm{Z}}_N$, which is also the output of Translator. To diversify the learned features, we adopt the multi-head attention mechanism, and group the features along the channel dimension. Then, the model learns the enhanced features by group and concatenates them along the channel dimension. Note that the channel dimension of the last PLA layer is changed to $(T^\prime \cdot \tilde{C})$ by $1\times 1$ convolution, where $T^\prime$ is the number of predicted frames.
  
\subsection{Decoder}
To decode the updated spatiotemporal features from Translator, we stack $M$ $3\times 3$ transposed 2D convolution layers $unConv(\cdot)$, each of which is followed by group normalization and leaky ReLU as an activation function. Among them, transposed convolution is used to up-sample the features at every two convolution layers, \ie, the stride is set to 2 and the size is enlarged from $(\frac{H}{2}, \frac{W}{2})$ to $(H, W)$. Its input is the enhanced spatiotemporal feature $\bar{\bm{Z}}_N\in \mathbb{R}^{B \times (T^\prime\cdot \tilde{C}) \times H^\prime \times W^\prime}$, whose size is reshaped to $(B \cdot T^\prime) \times \tilde{C} \times H^\prime \times W^\prime$. These are actually the features of predicted frames, and the features are passed through a series of group normalization and leaky ReLU units as well as a $1\times 1$ convolution layer, resulting in the predicted video sequence $\hat{\mathcal{Y}}_{T^\prime}\in \mathbb{R}^{B\times T^\prime\times C\times H\times W}$.

\subsection{Spatial Masking in Pretraining}
To strengthen the ability of the model to learn spatial features, we introduce the Spatial Masking (SM) strategy in pretraining. As illustrated in Fig.~\ref{fig:sm_module} (batch size $B$ and frame number $T$ omitted), Translator is substituted by the SM module, which adopts the attention mechanism with masking the feature map. Note that the input frames are randomly masked at the pixel level with some ratio $r_0>0$, and the masked frame is $\bar{\bm{X}}_t \in \mathbb{R}^{C\times H\times W}$. 

For $B$ batches with each batch containing $T$ frames, the input of SM module $SM(\cdot)$ is the feature $\hat{\bm{S}} \in \mathbb{R}^{(B\cdot T)\times \tilde{C}\times H^\prime\times W^\prime}$ derived from Encoder, \ie, $\hat{\bm{S}} = \mathcal{E}_{\Phi}(\bar{\mathbf{X}}_t)$, where $\mathcal{E}_{\Phi}(\cdot)$ is Encoder with its parameter $\Phi$. It can be thought as that the feature $\bm{S}$ is learned frame by frame. Here, we take one frame for example. Similar to PLA, we use global average pooling and $1\times 1$ convolution to project the spatial pixels ($\tilde{C}\times H^\prime \times W^\prime$) to one point ($\tilde{C}\times 1 \times 1$) along the channel dimension, resulting in \textit{Query} feature $\bm{Q}$ and \textit{Key} feature $\bm{K}$. Both of them are then squeezed to ($\tilde{C}\times 1$) by further reducing the dimension, \ie, $\{\bm{Q}, \bm{K}\}\in \mathbb{R}^{\tilde{C}\times 1}$. The two kinds of features are multiplied to derive the attention map $\bm{A}\in \mathbb{R}^{\tilde{C}\times \tilde{C}}$, \ie, $\bm{A} = \frac{\bm{Q}\bm{K}^\top}{\sqrt{\tilde{C}}}$. Given a mask ratio $0\le r\le 1$, we add the mask to the entries of the attention matrix $\bm{A}$ whose scores are the leading $\lfloor r\cdot \tilde{C}^2 \rfloor$ values in a descending order, where $\lfloor \cdot \rfloor$ denotes the fraction is rounded down. To achieve masking, those leading entries are set to $-\infty$ since they will turn to zeros by the $Softmax(\cdot)$ function, resulting in the masked attention matrix $\hat{\bm{A}}\in \mathbb{R}^{\tilde{C}\times \tilde{C}}$. Moreover, we feed the feature $\hat{\bm{S}}$ to the $3\times 3$ depth-wise convolution to obtain the \textit{Value} feature $\bm{V}\in \mathbb{R}^{\tilde{C}\times H^\prime \times W^\prime}$. Then, we compute the product of the masked attention matrix $\hat{\bm{A}}$ and the \textit{Value} feature $\bm{V}$, obtaining the masked spatial feature, \ie, $Softmax(\hat{\bm{A}})\cdot \bm{V} \in \mathbb{R}^{\tilde{C}\times H^\prime \times W^\prime}$. In this way, we repeat the above operations frame by frame in batch and concatenate these features, leading to the final masked spatial feature $\tilde{\bm{S}}\in \mathbb{R}^{(B\cdot T)\times \tilde{C}\times H^\prime \times W^\prime}$, which is then fed into Decoder to produce the reconstructed video sequence $\hat{\mathcal{X}}_{T}={\{\hat{\mathbf{X}}_t}\}_{t=1}^T$.

\subsection{Loss Function}
\textbf{Pretraining}. During pretraining, the model consists of Encoder, Spatial Masking module, and Decoder. The goal of the model is to minimize the empirical error between the source video frames and the frames derived from decoding the masked feature maps (reconstruction process). Specifically, the reconstruction loss $\mathcal{L}_{rec}$ adopts Mean Square Error (MSE), \ie,
\begin{equation}
	\label{eq:rec_loss}
	\mathcal{L}_{rec} = \min_{\Phi,\Omega}\sum_{t=1}^{T}
	\left \| \mathbf{X}_t-\mathcal{D}_{\Omega}(SM(\hat{\bm{S}})) \right \|^2_2,
\end{equation}
where $\|\cdot\|_2$ denotes $\ell_2$ norm, $\hat{\bm{S}}$ is the feature derived from Encoder, $\mathcal{D}_{\Omega}(\cdot)$ is Decoder with its parameter $\Omega$. Essentially, the reconstruction loss minimizes the error between the $t$-th frame $\bm{X}_t$ and its reconstructed frame $\hat{\mathbf{X}}_t =\mathcal{D}_{\Omega}(SM(\hat{\bm{S}}))$. In another word, it is expected that the reconstructed frame approaches the source frame where the spatial masking strategy is applied to.

\textbf{Training}. During training, the model consists of Encoder, Translator, and Decoder. The parameters in Encoder are frozen and those in Decoder are updated as Translator dynamically captures the spatiotemporal variations. Mathematically, we adopt MSE loss to minimize the empirical error between the source frames and the predicted frames, \ie, 
\begin{equation}
	\label{eq:train_loss}
	\mathcal{L} = \min\limits_{\Psi,\Omega}\sum_{t=T+1}^{T+T^\prime}
	\left \| \mathbf{X}_t-\mathcal{D}_{\Omega}(\mathcal{T}_{\Psi}(\mathcal{E}_{\Phi}(\mathbf{X}_{1:T}))_t) \right \|^2_2,
\end{equation}
where $\mathcal{T}_{\Psi}(\cdot)$ is Translator with its parameter $\Psi$ which outputs the updated spatiotemporal features, $T$ denotes the input frame number, and $T^\prime$ denotes the predicted frame number. Note that the parameter $\Phi$ in Encoder is fixed in training. In particular, each training clip contains $T+T^\prime$ frames, where the former $T$ frames are the model input and the latter $T^\prime$ frames are used for computing the MSE loss. 

%-------------------------------------------------------------------------
\section{Experiment}
\label{test}
All experiments were carried out on a machine with three NVIDIA RTX 3090 graphics cards, and the model was compiled using PyTorch 1.12, Python 3.10, and CUDA 11.1.

%-------------------------------------------------------------------------
\subsection{Datasets and Evaluation Metrics}
We comprehensively investigate the performance of several video prediction methods on five benchmarks, and show details in the following.

\textbf{Moving MNIST}\footnote{\url{https://www.cs.toronto.edu/~nitish/unsupervised_video/}}~\cite{srivastava-icml2015-movingmnist}. It consists of paired evolving hand-written digits from the MNIST\footnote{\url{http://yann.lecun.com/exdb/mnist/}} data set. Following \cite{wang-nips2017-predrnn}, training set has 10000 sequences and test set has 5000 sequences. Each sequence consists of 20 successive $64\times 64$ frames with two randomly appearing digits. Among them, 10 frames are the input and the rest are the output. The initial position and rate of each digit are random, but the rate keeps the same across the entire sequence.

\textbf{TaxiBJ}\footnote{\url{https://github.com/TolicWang/DeepST/tree/master/data/TaxiBJ}}~\cite{zhang-aaai2017-trafficbj}. It is collected from the real-world traffic scenario in Beijing, ranging from 2013 to 2016. The traffic flows have strong temporal dependency among nearby area, and the data pre-processing follows \cite{wang-cvpr2019-mim}. The data of the last four weeks are used as the test set (1334 clips) while the rest are the training set (19627 clips). Each clip has 8 frames, where 4 frames are the input and the others are the output. The size of each video frame is $32\times 32\times 2$, and the two channels indicate the in and the out traffic flow. 

\textbf{Human3.6M}\footnote{\url{http://vision.imar.ro/human3.6m/description.php}}~\cite{ionescu--tpami2014-human3.6m}. It contains the sports videos of 11 subjects in 17 scenes, involving 3.6 million human pose images from 4 distinct camera views. Following \cite{wang-cvpr2019-mim}, we use the data in the walking scene, which includes $128\times 128\times 3$ RGB frames. The subsets $\{S1, S5{\rm -}S8\}$ are for training (2624 clips) and $\{S9, S11\}$ are for test (1135 clips). Each clip has 8 frames, and the half of them are input. 

\textbf{KITTI\&Caltech Pedestrian}\footnote{\url{https://www.cvlibs.net/datasets/kitti/}}. Following \cite{wang-cvpr2019-mim}, we use 2042 clips in KITTI \cite{geiger-ijrr2013-kitti} for training and 1983 clips in Caltech Pedestrian \cite{dollar-cvpr2009-caltech} for test. Both of them are driving databases taken from a vehicle in an urban environment, and the RGB frames are resized to $128\times 160$ by center-cropping and downsampling. The former includes ``city'', ``residential'', and ``road'' categories, while the latter has about 10 hours of $640\times 480$ video. Each clip has 20 consecutive frames, where 10 frames are input and the others are output during training.

\textbf{KTH}\footnote{\url{https://www.csc.kth.se/cvap/actions/}}~\cite{schuldt-icpr2004-kth} includes six action classes, \ie, walking, jogging, running, boxing, hand waving, and hand clapping, involving 25 subjects in four different scenes. Each video clip is taken in 25 fps and is 4 seconds on average. Following \cite{villegas-iclr2017-mcnet}, the gray-scale frames are resized to $128\times 128$. The training set has 5200 clips (16 subjects) and the test set has 3167 clips (9 subjects). Each clip has 30 frames, where 10 frames are input and 20 frames are output during training.

\textbf{Evaluation Metrics}. Following \cite{guen-cvpr2020-phydnet}\cite{yu-iclr2020-crevnet}\cite{gao-cvpr2022-simvp}, we employ MSE (Mean Square Error), MAE (Mean Absolute Error), SSIM (Structure Similarity Index Measure) \cite{wang-tip2004-ssim}, and PSNR (Peak Signal to Noise Ratio) to evaluate the quality of the predicted frames. On Caltech Pedestrian \cite{dollar-cvpr2009-caltech}, we use MSE, SSIM, and PSNR; on KTH \cite{schuldt-icpr2004-kth}, we use SSIM and PSNR; on the remaining ones, we use MAE, MSE, and SSIM. SSIM ranges from -1 to 1, and the images are more similar when it approaches 1. The larger the PSNR db value, the better quality the video prediction model achieves.

% ------------- Parameter Setting ---------------
\begin{table}[!t]
	\centering
	\caption{Parameter setting.}
	\label{tbl:parameter}
	\setlength{\tabcolsep}{0.9mm}{ 
	\begin{tabular}{l c c c c c c c c c}
		\toprule[0.75pt]
		Dataset                                            &$(H,W,C)$  & ${T, T^\prime}$ & $r_0$  & $\tilde{C}$ & $\hat{C}$ & $M$ & $N$  & Epoch \\
		\midrule[0.5pt]
		Moving MNIST\cite{srivastava-icml2015-movingmnist} &(64,64,1)  & 10,10 &0.96          &   64  &  512  &  4  & 4 & 2000\\
		TaxiBJ\cite{zhang-aaai2017-trafficbj}              &(32,32,2)  & 4,4   &0.97          &   32  &  256  &  2  & 4 & 50\\
		Human3.6M\cite{ionescu--tpami2014-human3.6m}       &(128,128,3)& 4,4   &0.95          &   64  &  128  &  2  & 3 & 50\\
		KITTI\&Caltech\cite{geiger-ijrr2013-kitti}         &(128,160,3)& 10,1  &0.95          &   64  &  256  &  2  & 3 & 50\\
		KTH\cite{schuldt-icpr2004-kth}                     &(128,128,1)& 10,20/40 &0.90      &   32  &  128  &  3  & 3 & 100\\
		\toprule[0.75pt]
	\end{tabular}
}
\end{table}

\subsection{Experimental Settings}
\textbf{Pretraining Phase}. All parameters are initialized using the Kaiming initialization \cite{he-iccv2015-kaiming}, the learning rate is set to 0.01, and batch size $B$ is 16. The mask ratio $r$ in SM module is set to 0.1, and we use the Adam algorithm to pre-train the model by 50 epochs on all the data sets. 

\textbf{Training Phase}. For initialization, Encoder and Decoder use the parameters of pretraining, while Translator adopts the Kaiming initialization. We use the Adam algorithm to train the model with the momentum $\{\beta_1, \beta_2\} = \{0.9, 0.999\}$. The learning rate is set to 0.01 for MNIST \cite{srivastava-icml2015-movingmnist} and 0.001 for the rest, while it is adjusted by adopting cosine learning rate for TaxiBJ \cite{zhang-aaai2017-trafficbj} and one cycle learning rate \cite{simth-ispp2019-onecycle} for the rest. The head number in PLA module is set to 2 for Moving MNIST, Human3.6M, and KTH, while setting it to 8 for the rest. Moreover, the input frame size, input and output frame numbers $\{T, T^\prime\}$, pretraining mask ratio $r_0$, channel number $\tilde{C}$ of the spatial feature from Encoder and $\hat{C}$ of the spatiotemporal feature from Translator, number of Encoder and Decoder $M$, number of ConvNeXt and PLA blocks $N$, and the training epochs are recorded in Table~\ref{tbl:parameter}. Note that we use NNI (Neural Network Intelligence) tool\footnote{\url{https://nni.readthedocs.io/zh/stable/index.html}} to search for the optimal hyper-parameters $\tilde{C}$, $\hat{C}$, $M$, and $N$. In addition, we randomly mask the pixels of input frames at the ratio $r_0$ (pixel percentage) in pretraining.

\textbf{Test Phase}. We fix the model parameters of training model and feed the test video sequence to the model, which outputs $T^\prime$ predicted frames.

% --------- Results on Moving MNIST, TaxiBJ, and Huamn3.6M  -------
\begin{table*}[!t]
	\centering
	\caption{Comparison results on Moving MNIST \cite{srivastava-icml2015-movingmnist}, TaxiBJ \cite{zhang-aaai2017-trafficbj}, and Human3.6M \cite{ionescu--tpami2014-human3.6m}.}
	\label{tbl:mnist_taxibj_human}
	%\small
	\setlength{\tabcolsep}{1.4mm}{
		\begin{tabular}{lr ccc c lll c lll}
			\toprule[0.75pt]
			\multirow{2}{*}{Method} & \multirow{2}{*}{Venue} & \multicolumn{4}{c}{Moving MNIST\cite{srivastava-icml2015-movingmnist}} & \multicolumn{4}{c}{TaxiBJ\cite{zhang-aaai2017-trafficbj}} & \multicolumn{3}{c}{Human3.6M\cite{ionescu--tpami2014-human3.6m}}  \\ \cmidrule[0.5pt]{3-5}  \cmidrule[0.5pt]{7-9} \cmidrule[0.5pt]{11-13}
			& & MSE$\downarrow$ & MAE$\downarrow$ & SSIM$\uparrow$ & & MSE$\downarrow$ & MAE$\downarrow$ & SSIM$\uparrow$ && MSE$\downarrow$ & MAE$\downarrow$ &SSIM$\uparrow$  \\ \midrule[0.5pt]
			PredRNN\cite{wang-nips2017-predrnn}    & NeurIPS'17& 56.8 & 126.1 & 0.867 & & 46.4 & 17.1 & 0.971 && 48.4 & 18.9 & 0.781 \\
			PredRNN++\cite{wang-icml2018-predrnn++}& ICML'18   & 46.5 & 106.8 & 0.898 & & 44.8 & 16.9 & 0.977 && 45.8 & 17.2 & 0.851 \\
			MIM\cite{wang-cvpr2019-mim}            & CVPR'19   & 44.2 & 101.1 & 0.910 & & 42.9 & 16.6 & 0.971 && 42.9 & 17.8 & 0.790 \\
			E3D-LSTM\cite{wang-iclr2019-e3dlstm}   & ICLR'19   & 41.3 & 86.4  & 0.910 & & 43.2 & 16.9 & 0.979 && 46.4 & 16.6 & 0.869 \\
			\midrule[0.5pt]
			PhyDNet\cite{guen-cvpr2020-phydnet}    & CVPR'20   & 24.4 & 70.3  & 0.947 & & 41.9 & \underline{16.2} & \underline{0.982} && 36.9 & 16.2 & 0.901 \\
			CrevNet\cite{yu-iclr2020-crevnet}      & ICLR'20   & 22.3 & - & \underline{0.949} & & - & - & - && - & - & - \\
			MAU\cite{chang-nips2021-mau}            & NeurIPS'21& 27.6 & 80.3  & 0.937 & &42.2$^{\ast}$  & 16.4$^{\ast}$ & 0.982$^{\ast}$ && \underline{31.2}$^{\ast}$  & 15.0$^{\ast}$& 0.885$^{\ast}$ \\
			SimVP\cite{gao-cvpr2022-simvp}         & CVPR'22   & 23.8 & \underline{69.9} & 0.948 & & \underline{41.4} & \underline{16.2} & \underline{0.982} && 31.6 & 15.1 & \underline{0.904} \\
			PredRNNv2\cite{wang-tpami2023-predrnnv2}& TPAMI'23   & \underline{19.9} &  - & 0.939 & & 45.6$^{\ast}$ & 16.8$^{\ast}$ & 0.980$^{\ast}$ && 36.3$^{\ast}$ & 17.7$^{\ast}$ & 0.863$^{\ast}$ \\  
			STAM\cite{chang-tmm2022-stam}          & TMM'23    & 28.6 &   -   & 0.935 & & 44.1 &  -   &  -    &&  -  & \underline{13.2} & 0.875 \\ 
			\midrule[0.5pt]
			PLA-SM                                & Ours         & \textbf{18.4} & \textbf{57.6} & \textbf{0.960} & & \textbf{40.1} & \textbf{15.9} & \textbf{0.983} && \textbf{28.6} & \textbf{12.3} & \textbf{0.909} \\
			\toprule[0.75pt]
		\end{tabular}
	}
\end{table*}

% ------- Results on Caltech Pedestrian -----------
\begin{table}[!t]
	\centering
	\caption{Comparison results on Caltech Pedestrian \cite{dollar-cvpr2009-caltech}.}
	\label{tbl:caltech}
	\small
	\setlength{\tabcolsep}{1mm}{
		\begin{tabular}{l r c c c}
			\toprule[0.75pt]
			\multirow{2}{*}{Method} & \multirow{2}{*}{Venue} & \multicolumn{3}{c}{Caltech Pedestrian\cite{dollar-cvpr2009-caltech}(10$\rightarrow$1) } \\ \cmidrule[0.5pt]{3-5} 
			& &MSE$\downarrow$ & SSIM$\uparrow$ & PSNR$\uparrow$(db) \\
			\midrule[0.5pt]
			DVF\cite{liu-iccv2017-dvf}                  & ICCV'17   & -    & 0.897 & 26.2 \\
			Dual-GAN\cite{liang-iccv2017-dualmotiongan} & ICCV'17   & 2.41 & 0.899 & -    \\
			PredNet\cite{lotter-iclr2017-prednet}       & ICLR'17   & 2.42 & 0.905 & 27.6 \\
			CtrlGen\cite{hao-cvpr2018-ctrlgen}          & CVPR'18   & -    & 0.900 & 26.5 \\
			ContextVP\cite{byeon-eccv2018-contextvp}    & ECCV'18   & 1.94 & 0.921 & 28.7 \\
			DPG\cite{gao-iccv2019-dpg}                  & ICCV'19   & -    & 0.923 & 28.2 \\
			STMFANet\cite{jin-cvpr2020-stmfanet}        & CVPR'20   & 1.59 & 0.927 & 29.1 \\
			CrevNet\cite{yu-iclr2020-crevnet}           & ICLR'20   & 1.55 & 0.925 & 29.3 \\
			MAU\cite{chang-nips2021-mau}                & NeurIPS'21& 1.24 & 0.943 & 30.1 \\
			VPCL\cite{geng-cvpr2022-vpcl}               & CVPR'22   & -    & 0.928 & -    \\
			SimVP\cite{gao-cvpr2022-simvp}              & CVPR'22   & 1.56 & 0.940 & \underline{33.1} \\
			STAM\cite{chang-tmm2022-stam}               & TMM'23   & \underline{1.11} & \underline{0.945} & 29.9 \\
			\midrule[0.5pt]
			PLA-SM & Ours  & \textbf{1.09} & \textbf{0.953} & \textbf{33.7} \\
			\toprule[0.75pt]
		\end{tabular}
	}
\end{table}

% ------- Results on KTH -----------
\begin{table}[!t]
	\centering
	\caption{Comparison results on KTH \cite{schuldt-icpr2004-kth}.}
	\label{tbl:kth}
	\small
	\setlength{\tabcolsep}{0.85mm}{
		\begin{tabular}{l r c c c c}
			\toprule[0.75pt]
			\multirow{2}{*}{Method} & \multirow{2}{*}{Venue} & \multicolumn{2}{c}{KTH\cite{schuldt-icpr2004-kth}(10$\rightarrow$20)} & \multicolumn{2}{c}{KTH\cite{schuldt-icpr2004-kth}(10$\rightarrow$40)}\\ \cmidrule[0.5pt]{3-6} 
			& &SSIM$\uparrow$ & PSNR$\uparrow$(db) & SSIM$\uparrow$ & PSNR$\uparrow$(db) \\
			%\midrule[0.5pt]
			DFN\cite{jia-nips2016-dfn}             & \scriptsize{NeurIPS'16} & 0.794 & 27.26 & 0.652 & 23.01 \\
			MCnet\cite{villegas-iclr2017-mcnet}    & ICLR'17    & 0.804 & 25.95 & 0.730 & 23.89 \\
			PredRNN\cite{wang-nips2017-predrnn}    & \scriptsize{NeurIPS'17} & 0.839 & 27.55 & 0.703 & 24.16 \\
			fRNN\cite{oliu-eccv2018-frnn}          & ECCV'18    & 0.771 & 26.12 & 0.678 & 23.77 \\
			SV2P\cite{babaeizadeh-iclr2018-sv2p}   & ICLR'18    & 0.838 & 27.79 & 0.789 & 26.12 \\
			\scriptsize{PredRNN++}\cite{wang-icml2018-predrnn++}& ICML'18    & 0.865 & 28.47 & 0.741 & 25.21 \\
			SAVP\cite{lee-iclr2019-savp}           & ICLR'19    & 0.852 & 27.77 & 0.811 & 26.18 \\
			\scriptsize{E3D-LSTM}\cite{wang-iclr2019-e3dlstm}   & ICLR'19    & 0.879 & 29.31 & 0.810 & 27.24 \\
			\scriptsize{STMFANet}\cite{jin-cvpr2020-stmfanet}   & CVPR'20    & 0.893 & 29.85 & 0.851 & 27.56 \\
			GridVP\cite{gao-iros2021-gridkeypoint} & IROS'21    & -     & -     & 0.837 & 27.11 \\
			SimVP\cite{gao-cvpr2022-simvp}         & CVPR'22    & \underline{0.905} & \underline{33.72} & \underline{0.886} & \underline{32.93} \\
			\scriptsize{PredRNNv2}\cite{wang-tpami2023-predrnnv2}& TPAMI'23   & 0.838 & 28.37 &   -   &   -   \\
			\midrule[0.5pt]
			PLA-SM & Ours  & \textbf{0.909} & \textbf{34.31} & \textbf{0.888} & \textbf{33.41} \\
			\toprule[0.75pt]
		\end{tabular}
	}
\end{table}

\subsection{Compared Methods}
To examine the performance of the proposed PLA-SM approach, we compare nine SOTA alternatives on Moving MNIST \cite{srivastava-icml2015-movingmnist}, TaxiBJ \cite{zhang-aaai2017-trafficbj}, and Human3.6M \cite{zhang-aaai2017-trafficbj}. They involve PredRNN \cite{wang-nips2017-predrnn}, PredRNN++ \cite{wang-icml2018-predrnn++}, PredRNNv2 \cite{wang-tpami2023-predrnnv2}, MIM (Memory In Memory) \cite{wang-cvpr2019-mim}, E3D-LSTM \cite{wang-iclr2019-e3dlstm}, PhyDNet (Physical Dynamics Network) \cite{guen-cvpr2020-phydnet}, CrevNet \cite{yu-iclr2020-crevnet}, MAU (Motion-Aware Unit) \cite{chang-nips2021-mau}, SimVP \cite{gao-cvpr2022-simvp}, PredRNNv2 \cite{wang-tpami2023-predrnnv2}, and STAM (Spatio Temporal Attention Memory) \cite{chang-tmm2022-stam}. 

On Caltech Pedestrian \cite{dollar-cvpr2009-caltech}, we compare twelve SOTA methods, including DVF (Deep Voxel Flow) \cite{liu-iccv2017-dvf}, Dual-GAN (Dual Generative Adversarial Network) \cite{liang-iccv2017-dualmotiongan}, PredNet \cite{lotter-iclr2017-prednet}, CtrlGen (Controllable video Generation) \cite{hao-cvpr2018-ctrlgen}, ContextVP \cite{byeon-eccv2018-contextvp}, DPG (Disentangling Propagation and Generation) \cite{gao-iccv2019-dpg}, STMFANet (Spatial-Temporal Multi-Frequency Analysis Network) \cite{jin-cvpr2020-stmfanet}, CrevNet (Conditionally Reversible Network) \cite{yu-iclr2020-crevnet}, MAU (Motion-Aware Unit) \cite{chang-nips2021-mau}, VPCL (Video Prediction with Correspondence-wise Loss) \cite{geng-cvpr2022-vpcl}, SimVP \cite{gao-cvpr2022-simvp}, and STAM (Spatio-Temporal Attention Memory) \cite{chang-tmm2022-stam}.  

On KTH \cite{schuldt-icpr2004-kth}, we compare twelve competing algorithms, including DFN (Dynamic Filter Network) \cite{jia-nips2016-dfn}, MCNet \cite{villegas-iclr2017-mcnet}, PredRNN \cite{wang-nips2017-predrnn}, fRNN \cite{oliu-eccv2018-frnn}, SV2P (Stochastic Variational Video Prediction) \cite{babaeizadeh-iclr2018-sv2p}, PredRNN++ \cite{wang-icml2018-predrnn++}, SAVP (Stochastic Adversarial Video Prediction) \cite{lee-iclr2019-savp}, E3D-LSTM \cite{wang-iclr2019-e3dlstm}, STMFANet \cite{jin-cvpr2020-stmfanet}, GridVP \cite{gao-iros2021-gridkeypoint}, SimVP \cite{gao-cvpr2022-simvp}, and PredRNNv2 \cite{wang-tpami2023-predrnnv2}. 

% -------------------------------
\subsection{Quantitative Results}
We show the quantitative results of the compared methods on Moving MNIST \cite{srivastava-icml2015-movingmnist}, TaxiBJ \cite{zhang-aaai2017-trafficbj}, and Human3.6M \cite{ionescu--tpami2014-human3.6m} in Table~\ref{tbl:mnist_taxibj_human}.
The comparison results on Caltech Pedestrian \cite{dollar-cvpr2009-caltech} and KTH are respectively shown in Table~\ref{tbl:caltech} and Table~\ref{tbl:kth}. The best records are highlighted in bold and the second-best ones are underlined; ``-'' indicates the record is unavailable; ``*'' indicates the record is obtained by running the code provided by authors. 

\textbf{Moving MNIST, TaxiBJ, Human3.6M}. In Table~\ref{tbl:mnist_taxibj_human}, we follow \cite{gao-cvpr2022-simvp} to enlarge MSE values by 100 times for TaxiBJ, divide MSE and MAE values by 10 and 100 respectively for Human3.6M. From the table, we observe that the proposed PLA-SM method consistently outperforms the most competitive alternative across the three benchmark datasets. In particular, our approach improves the video prediction performance compared to the second best SimVP \cite{gao-cvpr2022-simvp} by 5.4, 12.3, and 1.1\% in terms of MSE, MAE, and SSIM, respectively. This demonstrates the designed pair-wise layer attention mechanism helps Translator to capture more accurate spatiotemporal dynamics in video sequence, which is beneficial for yielding higher-quality future frames. Previous methods such as PredRNN \cite{wang-nips2017-predrnn}, PredRNN++ \cite{wang-icml2018-predrnn++}, MIM\cite{wang-cvpr2019-mim}, and E3D-LSTM \cite{wang-iclr2019-e3dlstm} build the prediction model by stacking recurrent neural network blocks while the texture cues become very less when the depth increases a lot, leading to the obscure image. To alleviate this problem, some two-stage methods including PhyDNet \cite{guen-cvpr2020-phydnet}, CervNet \cite{yu-iclr2020-crevnet}, MAU \cite{chang-nips2021-mau}, and SimVP \cite{gao-cvpr2022-simvp} add skip connection between Encoder and Decoder, but it fails to make the texture feature be updated dynamically. Instead, our method not only adopts the spatial masking strategy in pretraining to obtain more robust encoding features, but also employs the stacked pair-wise layer attention blocks to generate the spatiotemporal features that contain more texture cues and high-level semantics simultaneously. 

\textbf{Caltech Pedestrian}. Different from other benchmarks, its training set and test set come from different sources, which facilitates evaluating the generalization ability of video prediction methods. As shown in Table~\ref{tbl:caltech}, our PLA-SM approach performs the best in terms of all three metrics, \ie, MSE, SSIM, and PSNR, when predicting one future frame given ten frames. This indicates the our method has the satisfying knowledge transfer ability in cross domain.

\textbf{KTH}. To examine the long-term prediction ability of our method, we use ten frames as input and evaluate the model performances when predicting 20 or 40 frames. As shown in Table~\ref{tbl:kth}, our method achieves better video prediction quality compared to those alternatives in both of two situations in terms of SSIM and PSNR. We attribute this to the fact that the spatial masking strategy in pretraining increases the ability of capturing the spatial structure, and the layer-wise attention mechanism makes the learned spatiotemporal feature involve more texture cues, which alleviates the obscure problem in some degree when the prediction sequence becomes longer.

% ------------ Computations on Moving MNIST -----------
\begin{table}[!t]
	\centering
	\caption{Computation comparison on Moving MNIST \cite{srivastava-icml2015-movingmnist}.}
	\label{tbl:computation}
	\small
	\setlength{\tabcolsep}{0.95mm}{
		\begin{tabular}{l r  r r c c c}
			\toprule[0.75pt]
			\multirow{2}{*}{Method}  & \multirow{2}{*}{Venue}  & FLOPs & Train & Test &\#Params   & MSE \\
			&   & (G)$\downarrow$ & (s)$\downarrow$ & (fps)$\uparrow$ & (M)$\downarrow$   &  $\downarrow$ \\
			\midrule[0.5pt]
			PredRNN\cite{wang-nips2017-predrnn}     & \scriptsize{NeurIPS}'17 & 115.6  & 300   & 107  & 23.8  &56.8 \\
			\scriptsize{PredRNN++}\cite{wang-icml2018-predrnn++} & ICML'18 & 171.7  & 530   & 70   & 38.6  &46.5 \\
			MIM\cite{wang-cvpr2019-mim}             & CVPR'19 & 179.2  & 564   & 55   & 38.0  &44.2 \\
			\scriptsize{E3D-LSTM}\cite{wang-iclr2019-e3dlstm}    & ICLR'19 & 298.9  & 1417  & 59   & 51.3  &41.3 \\
			\midrule[0.5pt]
			CrevNet\cite{yu-iclr2020-crevnet}       & ICLR'20 & 270.7  & 1030  & 10   & ~5.0   & \underline{22.3} \\
			PhyDNet\cite{guen-cvpr2020-phydnet}     & CVPR'20 & \underline{15.3}   & 196   & 63   & ~\textbf{3.1} &24.4 \\
			MAU\cite{chang-nips2021-mau}            & \scriptsize{NeurIPS}'21 & 17.8   & 210   & 58   & ~\underline{4.5}   &27.6 \\
			SimVP\cite{gao-cvpr2022-simvp}          & CVPR'22 & 19.4   & \underline{86}    & \underline{190}  &  22.3 &23.8 \\
			PredRNNv2\cite{wang-tpami2023-predrnnv2}& TPAMI'23 & 117.4  & 684   & 92   &  23.9 & \underline{19.9} \\
			\midrule[0.5pt]
			PLA-SM & Ours & \textbf{9.6} & \textbf{70} & \textbf{310} & 19.3 & \textbf{18.4} \\
			\toprule[0.75pt]
		\end{tabular}
	}
\end{table}

\textbf{Computational Analysis}. To show the efficiency of the proposed method, we compare nine alternatives in terms of FLOPs (G), training time (s), inference speed (fps), number of model parameters (M), and MSE on Moving MNIST, whose results are recorded in Table~\ref{tbl:computation}. Note that the top group methods adopt fixed training set, and the bottom group methods generate training samples online by randomly selecting two digits and motion path. The training time is computed for one epoch (per frame) using a single RTX3090 and the test time is the average fps of 10,000 samples. From the table, we see that those methods adopting recurrent neural networks or 3D convolutions require large computational costs, \eg, E3D-LSTM \cite{wang-iclr2019-e3dlstm} and CrevNet \cite{yu-iclr2020-crevnet} are the most two costly methods, which desire 1417 s and 1030 s per epoch, respectively, due to the expensive 3D convolutions. By the contrast, SimVP \cite{gao-cvpr2022-simvp} that adopts all convolutions strikes a good balance between the speed and the performance. Furthermore, we adopt the depth-wise convolutions in Translator, reducing FLOPs to 9.6G and speeding up the inference to 310 fps which are much better than the second-best candidate. This validates that our method enjoys better video prediction performance with lower computational cost and higher inference speed.

% -------------------------------
\subsection{Ablation Studies}
To probe into the inherent property of the proposed PLA-SM method, we do the ablations on the components, pretraining strategy, pretraining mask ratio $r_0$, the mask ratio $r$ of SM module, ConvNeXt module, and the head number of PLA module. Note that we vary the examined variable or component while the remaining settings keep still as in training unless specified.

% --------------- Ablations on Components ------------
\begin{table*}[!t]
	\centering
	\caption{Ablations on components of the proposed PLA-SM method. }
	\label{tbl:abl-component}
	\small
	\setlength{\tabcolsep}{1.0mm}{
		\begin{tabular}{ccc ccc c ccc c ccc c cc c cc}
			\toprule[0.75pt]
			\multicolumn{1}{l}{} & \multicolumn{1}{l}{} & \multicolumn{1}{l}{} & \multicolumn{4}{c}{Moving MNIST\cite{srivastava-icml2015-movingmnist}} & \multicolumn{4}{c}{TaxiBJ\cite{zhang-aaai2017-trafficbj}} & \multicolumn{4}{c}{Human3.6\cite{ionescu--tpami2014-human3.6m}} &\multicolumn{3}{c}{Caltech\cite{dollar-cvpr2009-caltech}} &\multicolumn{2}{c}{KTH\cite{schuldt-icpr2004-kth}} \\ 
			\cmidrule{4-6}  \cmidrule{8-10} \cmidrule{12-14} \cmidrule{16-17} \cmidrule{19-20}
			Mask & SM & PLA & MSE$\downarrow$ & MAE$\downarrow$ & SSIM$\uparrow$ & & MSE$\downarrow$ & MAE$\downarrow$ & SSIM$\uparrow$ & & MSE$\downarrow$ & MAE$\downarrow$ &SSIM$\uparrow$ && SSIM$\uparrow$ & PSNR$\uparrow$ &&SSIM$\uparrow$ & PSNR$\uparrow$\\  
			\midrule[0.5pt]
			&  &  & 22.8 & 62.4  & 0.952 && 43.3 & 16.2 & 0.981 && 31.2 & 15.8 & 0.901  && 0.944 & 32.09 &&0.904 & 33.42\\
			\checkmark &  &  & 21.6 & 60.8 & 0.956 && 42.2 & 16.1  & 0.981 && 30.0 & 14.5 & 0.904  && 0.946 & 32.35 && 0.906 & 33.85\\
			\checkmark & \checkmark &  & 20.1 & 59.5 & 0.958 && 41.0 & 15.9 & 0.982 && 29.0 & 12.7 & 0.907 && 0.948 & 32.70 && 0.908 & 34.12 \\
			&  & \checkmark & 21.2 & 61.2 & 0.956 && 42.3 & 16.1 & 0.982 && 30.6 & 14.8 & 0.903 && 0.948 & 32.85 && 0.906 & 33.97\\
			\checkmark & \checkmark & \checkmark & \textbf{18.4} & \textbf{57.6} & \textbf{0.960} && \textbf{40.1} & \textbf{15.9} & \textbf{0.983} &&  \textbf{28.6} & \textbf{12.3} & \textbf{0.909} && \textbf{0.953} & \textbf{33.72} && \textbf{0.909} & \textbf{34.31} \\
			\toprule[0.75pt]
		\end{tabular}
	}
\end{table*}

\textbf{Components}. We examined the effectiveness of the proposed Pair-wise Layer Attention (PLA) module, Spatial Masking (SM) module, and masking inputs on the five datasets. As shown in Table~\ref{tbl:abl-component}, the baseline without the three components (row~1) performs the worst across all evaluation metrics. When the masking strategy is applied to the input frames (row~2), the performance is boosted by 1.2\%, 1.1\%, and 1.2\% on Moving MNIST, TaxiBJ, and Human3.6M, respectively, in terms of MSE. This demonstrates that masking input frames is beneficial for learning better features to be decoded to future frames. When our spatial masking strategy is used in pretraining (row~3), the prediction performance is further improved by 1.3\% and 1.8\% on Moving MNIST and Human3.6M, respectively, in terms of MAE. This suggests that integrating the spatial masking with the attention mechanism helps to yield more robust features, whose Encoder parameters are used to initialize that in training. Meanwhile, when we just use the PLA module in Translator without pretraining (row~4), the performance is upgraded by 1.6\% and 1.0\% (compared to baseline), on Moving MNIST and TaxiBJ, respectively, in terms of MSE. This indicates that the stacked PLA blocks are good at capturing spatiotemporal dynamics that reflect the motion trend in the video sequence. Finally, when the pretraining strategy with spatial masking is employed together with PLA module (bottom row), the model achieves the most promising prediction performance, \eg, it strengthens the performance by 2.8\%, 2.2\%, and 2.0\% (compared to only using PLA) on Moving MNIST, TaxiBJ, and Human3.6M, respectively, in terms of MSE. This verifies that coupling the pretraining strategy with SM and PLA brings about the most benefits to the model for video prediction.

% ------------ Ablations on Pretraining -------------------
\begin{table*}[!t]
	\centering
	\caption{Ablations on the pretraining strategy.}
	\label{tbl:pretrain}
	\small
	\setlength{\tabcolsep}{0.6mm}{
		\begin{tabular}{lrc ccc c ccc c ccc c cc c cc}
			\toprule[0.75pt]
			\multicolumn{1}{l}{} & \multicolumn{1}{l}{} & \multicolumn{1}{l}{} & \multicolumn{4}{c}{Moving MNIST\cite{srivastava-icml2015-movingmnist}} & \multicolumn{4}{c}{TaxiBJ\cite{zhang-aaai2017-trafficbj}} & \multicolumn{4}{c}{Human3.6\cite{ionescu--tpami2014-human3.6m}} &\multicolumn{3}{c}{Caltech\cite{dollar-cvpr2009-caltech}} &\multicolumn{2}{c}{KTH\cite{schuldt-icpr2004-kth}} \\ \cmidrule{4-6}  \cmidrule{8-10} \cmidrule{12-14} \cmidrule{16-17} \cmidrule{19-20}
			Method & Venue & Pretrain & MSE$\downarrow$ & MAE$\downarrow$ & SSIM$\uparrow$ & & MSE$\downarrow$ & MAE$\downarrow$ & SSIM$\uparrow$ & & MSE$\downarrow$ & MAE$\downarrow$ &SSIM$\uparrow$ && SSIM$\uparrow$ & PSNR$\uparrow$ &&SSIM$\uparrow$ & PSNR$\uparrow$\\  \midrule[0.5pt]
			PhyDNet\cite{guen-cvpr2020-phydnet}  & CVPR'20 &  & 24.4 & 70.3 & 0.947 && 41.9 & 16.2 & 0.982 && 36.9 & 16.2 & 0.901  && - & - &&- & -\\
			&  & \checkmark & 23.2 & 67.2 & 0.949 && 41.6 & 16.1 & 0.982 && 33.4 & 15.8 & 0.902  && - & - && - & -\\
			MAU\cite{chang-nips2021-mau}  & \scriptsize{NeurIPS'21} &  & 27.6 & 80.3 & 0.937 && 42.2 & 16.4 & 0.982 && 31.2 & 15.0 & 0.885  && 0.943 & 30.12 &&- & -\\
			&  & \checkmark & 25.9 & 74.5 & 0.940 && 41.4 & 16.1 & 0.982 && 30.1 & 13.9 & 0.902  && 0.944 & 31.27 && - & -\\
			SimVP\cite{gao-cvpr2022-simvp} & CVPR'22 &  & 23.8 & 68.9 & 0.948 && 41.4 & 16.2 & 0.982 && 31.6 & 15.1 & 0.904  && 0.940 & 33.12 &&0.905 & 33.72\\
			&  & \checkmark & 22.3 & 66.4 & 0.950 && 40.8 & \textbf{15.9} & 0.982 && 29.8 & 13.4 & 0.906  && 0.945 & 33.64 &&0.907 & 34.14\\
			PLA-SM & Ours & \checkmark & \textbf{18.4} & \textbf{57.6} & \textbf{0.960} && \textbf{40.1} & \textbf{15.9} & \textbf{0.983} &&  \textbf{28.6} & \textbf{12.3} & \textbf{0.909} && \textbf{0.953} & \textbf{33.72} && \textbf{0.909} & \textbf{34.31} \\
			\toprule[0.75pt]			
		\end{tabular}
	}
\end{table*}

\textbf{Pretraining}. To investigate whether the pretraining strategy is also helpful for other methods, we show the results of several representative alternatives, including PhyDNet \cite{guen-cvpr2020-phydnet}, MAU \cite{chang-nips2021-mau}, and SimVP \cite{gao-cvpr2022-simvp}, in Table~\ref{tbl:pretrain}. From the table, we observe that the pretraining strategy improves the performance of all the methods on the five benchmarks, \eg, it reduces MSE by 1.2\%, 1.7\%, and 1.5\% for the three methods respectively. This demonstrates that masking input frames with spatial masking scheme indeed enhances the performance of other video prediction models. However, their performances are still inferior to ours, since we employ the pair-wise layer attention mechanism in Translator, which makes the model capture the motion trend better through spatiotemporal representations.

% --------------- Ablations on Pretraining Masking Ratio ----------
\begin{table*}[!t]
	\centering
	\caption{Ablations on the pretraining mask ratio $r_0$.}
	\label{tbl:abl-pretrainin-maskratio}
	\small
	\setlength{\tabcolsep}{1.3mm}{
		\begin{tabular}{c ccc c ccc c ccc c cc c cc}
			\toprule[0.75pt]
			\multirow{2}{*}{Mask ratio $r_0$} & \multicolumn{4}{c}{Moving MNIST\cite{srivastava-icml2015-movingmnist}} & \multicolumn{4}{c}{TaxiBJ\cite{zhang-aaai2017-trafficbj}} & \multicolumn{4}{c}{Human3.6\cite{ionescu--tpami2014-human3.6m}}& \multicolumn{3}{c}{Caltech\cite{dollar-cvpr2009-caltech}} & \multicolumn{2}{c}{KTH\cite{schuldt-icpr2004-kth}} \\ \cmidrule{2-4}  \cmidrule{6-8} \cmidrule{10-12} \cmidrule{14-15} \cmidrule{17-18}
			& MSE$\downarrow$ & MAE$\downarrow$ & SSIM$\uparrow$ & & MSE$\downarrow$ & MAE$\downarrow$ & SSIM$\uparrow$ && MSE$\downarrow$ & MAE$\downarrow$ &SSIM$\uparrow$ && SSIM$\uparrow$ & PSNR$\uparrow$ &&SSIM$\uparrow$ & PSNR$\uparrow$\\ \midrule[0.5pt]
			0.00    & 21.2 & 59.5 & 0.958 & & 42.2 & 16.2 & 0.981 && 29.8 & 13.6 & 0.903 && 0.948 & 32.87 && 0.906 & 33.88\\
			0.10    & 20.8 & 58.7 & 0.958 & & 42.0 & 16.2 & 0.981 && 29.6 & 13.6 & 0.903 && 0.949 & 32.89 && 0.907 & 33.94 \\
			0.30    & 20.5 & 58.3 & 0.958 & & 42.0 & 16.1 & 0.982 && 29.4 & 13.1 & 0.906 && 0.949 & 32.95 && 0.907 & 34.00\\
			0.50    & 20.2 & 58.2 & 0.959 & & 42.0 & 16.1 & 0.981 && 29.4 & 13.1 & 0.906 && 0.950 & 33.08 && 0.908 & 34.17 \\
			0.70    & 20.2 & 58.2 & 0.959 & & 41.6 & 16.0 & 0.982 && 29.1 & 12.8 & 0.907 && 0.950 & 33.16 && 0.908 & 34.22\\
			0.90    & 19.5 & 57.9  & 0.959 & & 41.2 & 16.0 & 0.982 && 28.7 & 12.6 & 0.908 && 0.953 & 33.67 && \textbf{0.909} & \textbf{34.31}\\
			\midrule[0.5pt]
			0.95    & 19.1 & 57.9  & 0.959 & & 41.2 & 15.9 & 0.982 && \textbf{28.6} & \textbf{12.3} & \textbf{0.909} && \textbf{0.953} & \textbf{33.72} &&0.905 & 33.46\\
			0.96    & \textbf{18.4} & \textbf{57.6} & \textbf{0.960} & & 40.9 & 15.9 & 0.983 && 28.7 & 12.5 & 0.908 && 0.952 & 33.67 &&0.905 & 33.21\\
			0.97    & 18.5 & 58.1  & 0.959 & & \textbf{40.1} & \textbf{15.9} & \textbf{0.983} && 29.1 & 12.7 & 0.907 && 0.950 & 32.89 &&0.904 & 33.02\\
			0.98    & 20.8 & 58.7 & 0.958 & & 41.1 & 16.0 & 0.983 && 29.6 & 12.8 & 0.906 && 0.947 & 32.65 && 0.903 & 32.86\\
			0.99    & 21.1 & 59.2 & 0.958 & & 43.3 & 16.2 & 0.982 && 30.0 & 13.4 & 0.905 && 0.947 & 32.42 && 0.900 & 31.92\\
			1.00    & 22.8 & 68.3 & 0.948 & & 43.6 & 16.4 & 0.982 && 32.2 & 13.9 & 0.902 && 0.945 & 32.12 && 0.898 & 31.27\\
			\toprule[0.75pt]
		\end{tabular}
	}
\end{table*}

\textbf{Pretraining mask ratio $r_0$}. During pretraining, we vary the mask ratio $r_0$ of input frames from 0 to 1.0 with a larger gap (0.1) before 0.9 and a smaller gap (0.01) after 0.95, whose results are recorded in Table~\ref{tbl:abl-pretrainin-maskratio}. As shown in the table, the MSE and the MAE values first decrease, and then increase in the range of 0 and 1. The prediction performance achieves the best when the masking area of a frame is over 90\% across all the five datasets, which indicates the larger mask ratio makes the Encoder be learned better during pretraining.

% -------------- Ablations on Mask Ratio in SM module --------------
\begin{table*}[!t]
	\centering
	\caption{Ablations on the mask ratio $r$ of the SM module without PLA module.}
	\label{tbl:abl-sm-maskratio}
	\small
	\setlength{\tabcolsep}{1.3mm}{
		\begin{tabular}{c ccc c ccc c ccc c cc c cc}
			\toprule[0.75pt]
			\multirow{2}{*}{Mask ratio $r$} & \multicolumn{4}{c}{Moving MNIST\cite{srivastava-icml2015-movingmnist}} & \multicolumn{4}{c}{TaxiBJ\cite{zhang-aaai2017-trafficbj}} & \multicolumn{4}{c}{Human3.6\cite{ionescu--tpami2014-human3.6m}} & \multicolumn{3}{c}{Caltech\cite{dollar-cvpr2009-caltech}} &\multicolumn{2}{c}{KTH\cite{schuldt-icpr2004-kth}}\\ \cmidrule{2-4}  \cmidrule{6-8}   \cmidrule{10-12}  \cmidrule{14-15}  \cmidrule{17-18}
			& MSE$\downarrow$ & MAE$\downarrow$ & SSIM$\uparrow$ & & MSE$\downarrow$ & MAE$\downarrow$ & SSIM$\uparrow$ && MSE$\downarrow$ & MAE$\downarrow$ &SSIM$\uparrow$ && SSIM$\uparrow$ & PSNR$\uparrow$ && SSIM$\uparrow$ & PSNR$\uparrow$\\ \midrule[0.5pt]
			0.00    & 22.8 & 62.4 & 0.952 & & 43.3 & 16.2 & 0.981 && 31.2 & 15.8 & 0.901 &&0.944 & 32.09 && 0.904 & 33.42\\
			0.05    & 22.2 & 61.8 & 0.954 & & 42.7 & 16.1 & 0.981 && 30.6 & 15.2 & 0.903 &&0.945 & 32.30 && 0.905 & 33.59\\
			0.10    & \textbf{21.5} & \textbf{60.4} & \textbf{0.956} & & \textbf{42.0} & \textbf{16.1} & \textbf{0.982} && \textbf{30.1} & \textbf{14.7} & \textbf{0.904} && \textbf{0.946} & \textbf{32.33} && \textbf{0.906} & \textbf{33.87}\\
			0.15    & 23.4 & 63.6 & 0.950 & & 42.9 & 16.3 & 0.981 && 31.4 & 16.1 & 0.899&& 0.943 & 31.97 && 0.904 & 33.35 \\
			\toprule[0.75pt]
		\end{tabular}
	}
\end{table*}

\textbf{SM mask ratio $r$}. For the Spatial Masking module, we vary the mask ratio $r$ of feature maps from 0 to 0.15 with an interval of 0.05, and show the results in Table~\ref{tbl:abl-sm-maskratio}. As we see from the table, our method consistently performs the best when the feature mask ratio is set to 0.1 across all the datasets. This indicates that it is sufficient to only mask a small proportion of feature map pixels, so as to help the model to learn the promising spatiotemporal features for decoding into future frames.

% -------------- ConvNeXt vs Vanilla Conv ---------------
\begin{table*}[!t]
	\centering
	\caption{Ablations on ConvNeXt module in Translator without pretraining.}
	\label{tbl:abl-convnext}
	\small
	\setlength{\tabcolsep}{1.2mm}{
		\begin{tabular}{cc ccc c ccc c ccc c cc c cc}
			\toprule[0.75pt]
			\multicolumn{1}{l}{} & \multicolumn{1}{l}{} & \multicolumn{4}{c}{Moving MNIST\cite{srivastava-icml2015-movingmnist}} & \multicolumn{4}{c}{TaxiBJ\cite{zhang-aaai2017-trafficbj}} & \multicolumn{4}{c}{Human3.6\cite{ionescu--tpami2014-human3.6m}} &\multicolumn{3}{c}{Caltech\cite{dollar-cvpr2009-caltech}} &\multicolumn{2}{c}{KTH\cite{schuldt-icpr2004-kth}} \\ \cmidrule{3-5}  \cmidrule{7-9} \cmidrule{11-13} \cmidrule{15-16} \cmidrule{18-19}
			ConvNeXt & PLA & MSE$\downarrow$ & MAE$\downarrow$ & SSIM$\uparrow$ & & MSE$\downarrow$ & MAE$\downarrow$ & SSIM$\uparrow$ & & MSE$\downarrow$ & MAE$\downarrow$ &SSIM$\uparrow$ && SSIM$\uparrow$ & PSNR$\uparrow$ &&SSIM$\uparrow$ & PSNR$\uparrow$\\  \midrule[0.5pt]
			&          & 28.3 & 82.2 & 0.935 && 45.6 & 16.8 & 0.980 && 34.7 & 17.6 & 0.894  && 0.936 & 31.04 && 0.898 & 32.74\\
			&\checkmark& 24.6 & 72.8 & 0.946 && 44.2 & 16.4 & 0.981 && 32.6 & 16.4 & 0.899  && 0.941 & 31.64 && 0.901 & 33.22\\
			\checkmark &          & 22.8 & 62.4 & 0.954 && 43.3 & 16.2 & 0.981 && 31.2 & 15.8 & 0.901  && 0.944 & 32.09 && 0.904 & 33.42\\
			\checkmark &\checkmark& \textbf{21.2} & \textbf{61.2} & \textbf{0.956} && \textbf{42.3} & \textbf{16.1} & \textbf{0.982} && \textbf{30.6} & \textbf{14.8} & \textbf{0.903}  && \textbf{0.948} & \textbf{32.85} && \textbf{0.906} & \textbf{33.97} \\
			\toprule[0.75pt]
		\end{tabular}
	}
\end{table*}

% ---------- Ablations on head number in Multi-head Attention ---------
\begin{table*}[!t]
	\centering
	\caption{Ablations on head number of multi-head attention in the PLA module.}
	\label{tbl:abl-headnum}
	\small
	\setlength{\tabcolsep}{1.3mm}{
		\begin{tabular}{c ccc c ccc c ccc c cc c cc}
			\toprule[0.75pt]
			\multirow{2}{*}{Head number} & \multicolumn{4}{c}{Moving MNIST\cite{srivastava-icml2015-movingmnist}} & \multicolumn{4}{c}{TaxiBJ\cite{zhang-aaai2017-trafficbj}} & \multicolumn{4}{c}{Human3.6\cite{ionescu--tpami2014-human3.6m}} &\multicolumn{3}{c}{Caltech\cite{dollar-cvpr2009-caltech}} &\multicolumn{2}{c}{KTH\cite{schuldt-icpr2004-kth}}\\ \cmidrule{2-4}  \cmidrule{6-8}  \cmidrule{10-12} \cmidrule{14-15}  \cmidrule{17-18}
			& MSE$\downarrow$ & MAE$\downarrow$ & SSIM$\uparrow$ & & MSE$\downarrow$ & MAE$\downarrow$ & SSIM$\uparrow$ && MSE$\downarrow$ & MAE$\downarrow$ &SSIM$\uparrow$ && SSIM$\uparrow$ & PSNR$\uparrow$ && SSIM$\uparrow$ & PSNR$\uparrow$ \\ \midrule[0.5pt]
			1  & 18.6 & 58.2 & 0.959 & & 41.2 & 16.0 & 0.983 && 28.7 & 12.7 & 0.907 && 0.953 & 33.69 && 0.908 & 34.16\\
			2  & \textbf{18.4} & \textbf{57.6} & \textbf{0.960} && 41.2 & 16.0 & 0.983 && \textbf{28.6} & \textbf{12.3} & \textbf{0.909} &&0.953 & 33.65 && \textbf{0.909} & \textbf{34.31}\\
			4  & 18.7 & 58.4 & 0.958 & & 40.6 & 16.0 & 0.983 && 28.7 & 12.7 & 0.907 && 0.953 & 33.70 && 0.909 & 34.27\\
			8  & 18.6 & 58.2 & 0.959 & & \textbf{40.1} & \textbf{15.9} & \textbf{0.983} && 28.6 & 12.5 & 0.908 && \textbf{0.953} & \textbf{33.72} && 0.908 & 34.19\\
			16 & 18.5 & 57.9 & 0.959 & & 40.2 & 16.2 & 0.982 && 28.7 & 12.6 & 0.908 && 0.953 & 33.68 && 0.908 & 34.22\\
			\toprule[0.75pt]	
		\end{tabular}
	}
\end{table*}

\textbf{ConvNeXt module}. As mentioned earlier, we stack the ConvNeXt blocks and the PLA blocks to learn the spatiotemporal features. Here we show their ablation results without the pretraining strategy in Table~\ref{tbl:abl-convnext}. Compared with the baseline which adopts $7\times 7$ vanilla convolution to substitute ConvNeXt (row~1), both PLA module and ConvNeXt module bring about the performance gains independently on all the five benchmarks. For example, PLA module reduces the MSE value by 3.7\%, 1.4\%, and 2.1\% on Moving MNIST, TaxiBJ, and Human3.6M, respectively, while it upgrades the performance by 0.6\% and 0.5\% on Caltech Pedestrian and KTH, respectively, in terms of PSNR. Similar performance improvements are observed when applying ConvNeXt to Translator rather than using vanilla convolution. Of course, the prediction performances are further enhanced by combining the two components together in Translator to model the motion dynamics of video.

\textbf{PLA head number}. To examine the influences of the head number in multi-head attention adopted by our PLA module, we vary the head number from 1 to 16 and show the results in Table~\ref{tbl:abl-headnum}. From the table, it can be observed that the prediction performance is robust to different head numbers in PLA module.

%% ---------- Ablations on Sparse Convolution -----------------
%\begin{table*}[!t]
%	\centering
%	\caption{The ablation study of sparse convolution on five datasets test set.}
%	\label{tbl:abl-sparseconv}
%	\small
%	\setlength{\tabcolsep}{1.2mm}{
%	\begin{tabular}{c ccc c ccc c ccc c cc c cc}
%		\toprule[0.75pt]
%		\multirow{2}{*}{Sparse Conv} & \multicolumn{4}{c}{Moving MNIST\cite{srivastava-icml2015-movingmnist}} & \multicolumn{4}{c}{TaxiBJ\cite{zhang-aaai2017-trafficbj}} & \multicolumn{3}{c}{Human3.6\cite{ionescu--tpami2014-human3.6m}} &\multicolumn{3}{c}{Caltech\cite{dollar-cvpr2009-caltech}} & \multicolumn{3}{c}{KTH\cite{schuldt-icpr2004-kth}}\\ \cmidrule{2-4}  \cmidrule{6-8}  \cmidrule{10-12}  \cmidrule{14-15} \cmidrule{17-18}
%		& MSE$\downarrow$ & MAE$\downarrow$ & SSIM$\uparrow$ & & MSE$\downarrow$ & MAE$\downarrow$ & SSIM$\uparrow$ && MSE$\downarrow$ & MAE$\downarrow$ &SSIM$\uparrow$ &&SSIM$\uparrow$ & PSNR$\uparrow$ && SSIM$\uparrow$ & PSNR$\uparrow$\\ \midrule[0.5pt]
%		& 24.2 & 70.1 & 0.947 & & 43.8 & 16.4 & 0.980 && 33.2 & 16.4 & 0.898 && 0.942 & 31.67 && 0.903 & 33.28\\
%		\checkmark    & 21.6 & 60.8 & 0.956 & & 42.2 & 16.1 & 0.982 && 30.0 & 14.5 & 0.904 && 0.946 & 32.35 &&0.906 & 33.85\\
%		\toprule[0.75pt]
%	\end{tabular}
%	}
%\end{table*}

\subsection{Qualitative Results}
\begin{figure}[!t]
	\centering
	\includegraphics[width=0.48\textwidth]{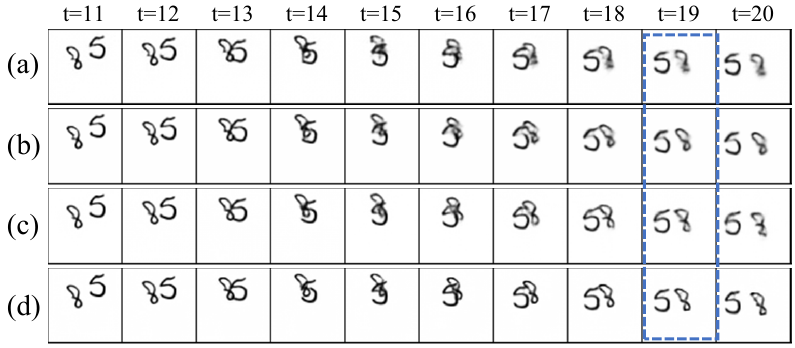}
	\caption{Predictions on Moving MNIST \cite{srivastava-icml2015-movingmnist}. (a) MAU \cite{chang-nips2021-mau}; (b) PhyDNet \cite{guen-cvpr2020-phydnet}; (c) SimVP \cite{gao-cvpr2022-simvp}; (d) Ours.}
	\label{fig:mnist}
\end{figure}

\begin{figure}[!t]
	\centering
	\includegraphics[width=0.48\textwidth]{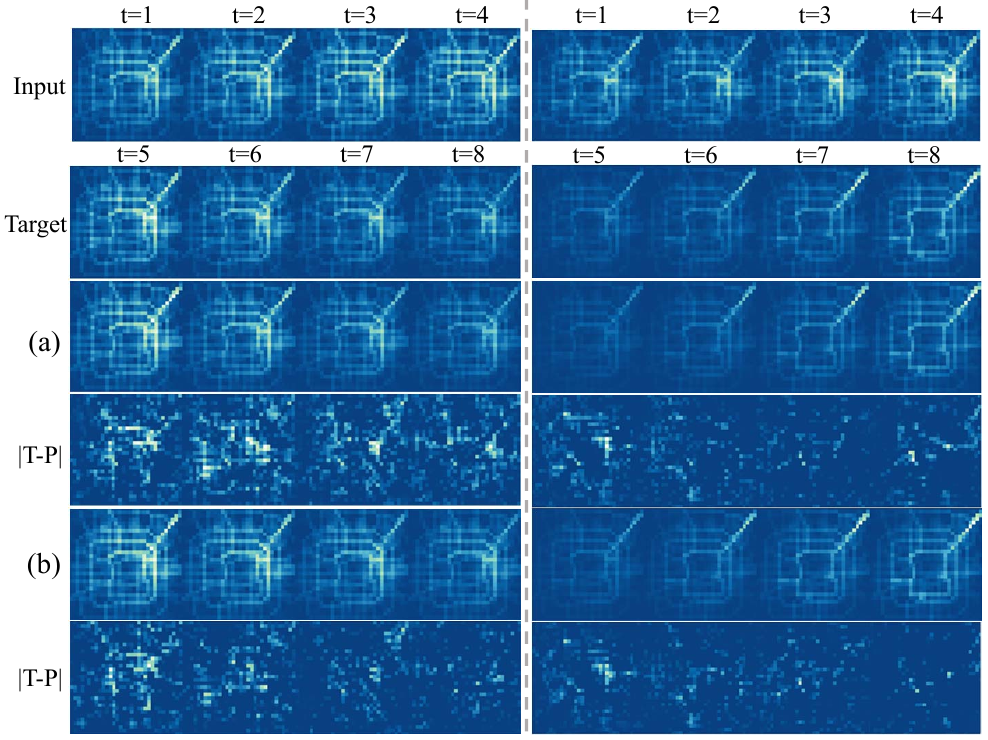}
	\caption{Predictions on TaxiBJ \cite{zhang-aaai2017-trafficbj}. (a) PredRNNv2 \cite{wang-tpami2023-predrnnv2}; (b) Ours.}
	\label{fig:taxibj}
\end{figure}

\begin{figure}[!t]
	\centering
	\includegraphics[width=0.42\textwidth]{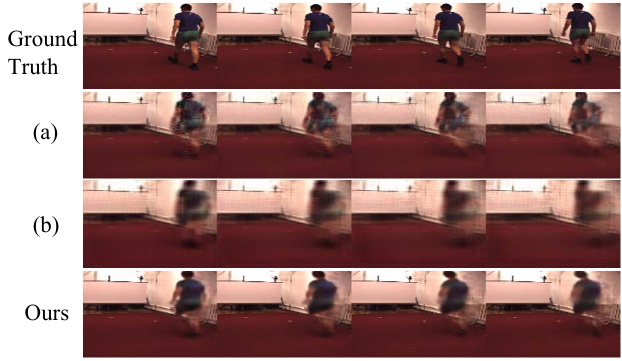}
	\caption{Predictions on Human3.6M \cite{ionescu--tpami2014-human3.6m}. (a) MIM \cite{wang-cvpr2019-mim}; (b) MAU \cite{chang-nips2021-mau}.}
	\label{fig:human3.6m}
\end{figure}

\begin{figure}[!t]
	\centering
	\includegraphics[width=0.45\textwidth]{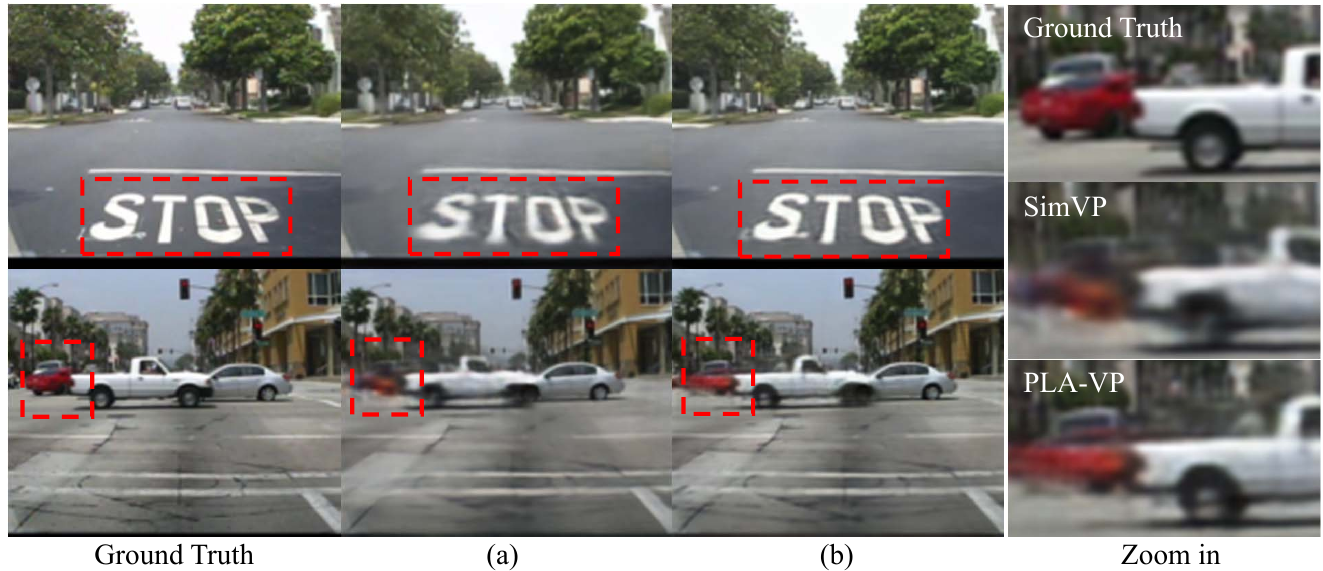}
	\caption{Predictions on Caltech Pedestrian \cite{dollar-cvpr2009-caltech}. (a) SimVP \cite{gao-cvpr2022-simvp}; (b) Ours.}
	\label{fig:caltech}
\end{figure}

\begin{figure}[!t]
	\centering
	\includegraphics[width=0.48\textwidth]{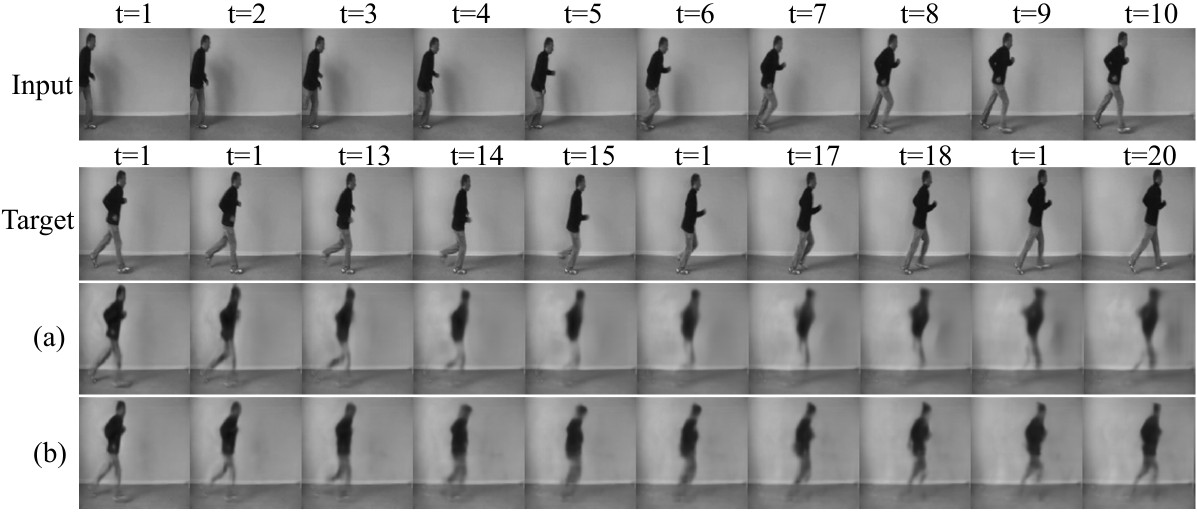}
	\caption{Predictions on KTH \cite{schuldt-icpr2004-kth}. (a) SimVP \cite{gao-cvpr2022-simvp}; (b) Ours.}
	\label{fig:kth}
\end{figure}

To intuitively visualize the video prediction results, we randomly choose some samples from the benchmarks and show the results in Fig.~\ref{fig:mnist} (Moving MNIST \cite{srivastava-icml2015-movingmnist}), Fig.~\ref{fig:taxibj} (TaxiBj \cite{zhang-aaai2017-trafficbj}), Fig.~\ref{fig:human3.6m} (Human3.6M \cite{ionescu--tpami2014-human3.6m}), Fig.~\ref{fig:caltech} (Caltech Pedestrian \cite{dollar-cvpr2009-caltech}), and Fig.~\ref{fig:kth} (KTH \cite{schuldt-icpr2004-kth}). 

From these figures, we have the following observations. 1) Fig.~\ref{fig:mnist} shows a predicted sequence with two digits `8' and `5' from $t=11$ to $t=20$, which indicates that our PLA-SM method (bottom row) generates clearer digits compared to other alternatives, including MAU \cite{chang-nips2021-mau} (top row), PhyDNet \cite{guen-cvpr2020-phydnet} (row~2), and SimVP \cite{gao-cvpr2022-simvp} (row~3). 2) Fig.~\ref{fig:taxibj} shows the traffic flow prediction frames and the difference map (row~4 / row~6) between predicted frame and ground-truth frame, which clearly displays that our prediction frames are more faithful to the ground-truth ones. 3) From Fig.~\ref{fig:human3.6m} to Fig.~\ref{fig:kth}, we see that the predictions of our method have less obscured areas compared to others, which well verifies the fact that the pre-training strategy with spatial masking helps to generate more robust encoding features, while the PLA module is good at capturing the motion trend in video.

% --- Visualization on Moving MNIST ------

% ------------- Visualization on TaxiBJ -------

% ------------- Visualization on Human3.6M -------

% ------------- Visualization on Caltech Pedestrian -------

% ------------- Visualization on KTH -------

\section{Conclusion}
\label{conclusion}
This work considers the video prediction problem from two perspectives, \ie, pretraining and utilizing low-level texture cues. In particular, we not only mask the input frames but also mask the feature maps by the proposed Spatial Masking module during pretraining, so as to make the learned encoding features be more robust. Meanwhile, we develop the Pair-wise Layer Attention mechanism for Translator by adopting the U-shape structure with ConvNeXt blocks, in order to employ the low-level cues to make compensations for the high-level features by skip connections. Such design allows the model to better capture the spatiotemporal dynamics which are of vital importance to yielding high-quality future frames. Comprehensive experiments in addition to rich ablations on five benchmarks have verified the effectiveness of the proposed PLA-SM approach both quantitatively and qualitatively.

However, there are still some limitations in our method. First, the mask ratio is fixed during pretraining, which makes it be insufficient to further increase the robustness of encoding features. It deserves the exploration of adaptively masking the feature maps to increase the feature diversity. Second, we adopt the symmetric U-shape structure to make interactions between the corresponding layers, which may neglect the different contributions of low-level and high-level representations made to the video prediction model. In future, it is interesting to investigate the asymmetry structure in Translator.

%\bibliographystyle{IEEEtran}
%\bibliography{plasm_vidpre}

\begin{thebibliography}{10}
	\providecommand{\url}[1]{#1}
	\csname url@samestyle\endcsname
	\providecommand{\newblock}{\relax}
	\providecommand{\bibinfo}[2]{#2}
	\providecommand{\BIBentrySTDinterwordspacing}{\spaceskip=0pt\relax}
	\providecommand{\BIBentryALTinterwordstretchfactor}{4}
	\providecommand{\BIBentryALTinterwordspacing}{\spaceskip=\fontdimen2\font plus
		\BIBentryALTinterwordstretchfactor\fontdimen3\font minus
		\fontdimen4\font\relax}
	\providecommand{\BIBforeignlanguage}[2]{{%
			\expandafter\ifx\csname l@#1\endcsname\relax
			\typeout{** WARNING: IEEEtran.bst: No hyphenation pattern has been}%
			\typeout{** loaded for the language `#1'. Using the pattern for}%
			\typeout{** the default language instead.}%
			\else
			\language=\csname l@#1\endcsname
			\fi
			#2}}
	\providecommand{\BIBdecl}{\relax}
	\BIBdecl
	
	\bibitem{li-tmm2023-ffinet}
	P.~Li, C.~Zhang, and X.~Xu, ``Fast fourier inception networks for occluded
	video prediction,'' \emph{IEEE Transactions on Multimedia (TMM)}, pp. 1--12,
	2023.
	
	\bibitem{yu-nc2019-rnn}
	Y.~Yu, X.~Si, C.~Hu, and J.~Zhang, ``A review of recurrent neural networks:
	Lstm cells and network architectures,'' \emph{Neural Computation}, vol.~31,
	no.~7, p. 1235–1270, 2019.
	
	\bibitem{wang-nips2017-predrnn}
	Y.~Wang, M.~Long, J.~Wang, Z.~Gao, and P.~S. Yu, ``Predrnn: Recurrent neural
	networks for predictive learning using spatiotemporal lstms,'' in
	\emph{Advances in Neural Information Processing Systems (NeurIPS)}, 2017, pp.
	879--888.
	
	\bibitem{wang-icml2018-predrnn++}
	Y.~Wang, Z.~Gao, M.~Long, J.~Wang, and P.~S. Yu, ``Predrnn++: Towards {a}
	resolution of the deep-in-time dilemma in spatiotemporal predictive
	learning,'' in \emph{Proceedings of the International Conference on Machine
		Learning (ICML)}, 2018, pp. 5110--5119.
	
	\bibitem{wang-cvpr2019-mim}
	Y.~Wang, J.~Zhang, H.~Zhu, M.~Long, J.~Wang, and P.~S. Yu, ``Memory in memory:
	{A} predictive neural network for learning higher-order non-stationarity from
	spatiotemporal dynamics,'' in \emph{Proceedings of the IEEE Conference on
		Computer Vision and Pattern Recognition (CVPR)}, 2019, pp. 9154--9162.
	
	\bibitem{wu-cvpr2021-motionrnn}
	H.~Wu, Z.~Yao, J.~Wang, and M.~Long, ``Motionrnn: {A} flexible model for video
	prediction with spacetime-varying motions,'' in \emph{Proceedings of the IEEE
		Conference on Computer Vision and Pattern Recognition (CVPR)}, 2021, pp.
	15\,435--15\,444.
	
	\bibitem{wang-iclr2019-e3dlstm}
	Y.~Wang, L.~Jiang, M.~Yang, L.~Li, M.~Long, and L.~Fei{-}Fei, ``Eidetic 3d
	{LSTM:} {A} model for video prediction and beyond,'' in \emph{Proceedings of
		the International Conference on Learning Representations (ICLR)}, 2019.
	
	\bibitem{yu-iclr2020-crevnet}
	W.~Yu, Y.~Lu, S.~Easterbrook, and S.~Fidler, ``Efficient and
	information-preserving future frame prediction and beyond,'' in
	\emph{Proceedings of the International Conference on Learning Representations
		(ICLR)}, 2020.
	
	\bibitem{chang-nips2021-mau}
	Z.~Chang, X.~Zhang, S.~Wang, S.~Ma, Y.~Ye, X.~Xiang, and W.~Gao, ``Mau: A
	motion-aware unit for video prediction and beyond,'' in \emph{Advances in
		Neural Information Processing Systems (NeurIPS)}, vol.~34, 2021.
	
	\bibitem{gao-cvpr2022-simvp}
	Z.~Gao, C.~Tan, L.~Wu, and S.~Z. Li, ``Simvp: Simpler yet better video
	prediction,'' in \emph{Proceedings of the IEEE Conference on Computer Vision
		and Pattern Recognition (CVPR)}, 2022, pp. 3160--3170.
	
	\bibitem{li-tnnls2022-cnn}
	Z.~Li, F.~Liu, W.~Yang, S.~Peng, and J.~Zhou, ``A survey of convolutional
	neural networks: analysis, applications, and prospects,'' \emph{IEEE
		Transactions on Neural Networks and Learning Systems (TNNLS)}, vol.~33,
	no.~12, pp. 6999--7019, 2022.
	
	\bibitem{li-pr2023-tad}
	P.~Li, J.~Cao, L.~Yuan, Q.~Ye, and X.~Xu, ``Truncated attention-aware proposal
	networks with multi-scale dilation for temporal action detection,''
	\emph{Pattern Recognition (PR)}, vol. 142, p. 109684, 2023.
	
	\bibitem{chang-cvpr2022-strpm}
	Z.~Chang, X.~Zhang, S.~Wang, S.~Ma, and W.~Gao, ``{STRPM:} {A} spatiotemporal
	residual predictive model for high-resolution video prediction,'' in
	\emph{Proceedings of the IEEE Conference on Computer Vision and Pattern
		Recognition (CVPR)}, 2022, pp. 13\,926--13\,935.
	
	\bibitem{guen-cvpr2020-phydnet}
	V.~L. Guen and N.~Thome, ``Disentangling physical dynamics from unknown factors
	for unsupervised video prediction,'' in \emph{Proceedings of the IEEE
		Conference on Computer Vision and Pattern Recognition (CVPR)}, 2020, pp.
	11\,474--11\,484.
	
	\bibitem{he-cvpr2022-mae}
	K.~He, X.~Chen, S.~Xie, Y.~Li, P.~Doll{\'{a}}r, and R.~B. Girshick, ``Masked
	autoencoders are scalable vision learners,'' in \emph{Proceedings of the IEEE
		Conference on Computer Vision and Pattern Recognition (CVPR)}, 2022, pp.
	15\,979--15\,988.
	
	\bibitem{jing-arxiv2022-mscn}
	L.~Jing, J.~Zhu, and Y.~LeCun, ``Masked siamese convnets,'' \emph{arXiv
		preprint arXiv:2206.07700}, 2022.
	
	\bibitem{liu-cvpr2015-sparseconv}
	B.~Liu, M.~Wang, H.~Foroosh, M.~F. Tappen, and M.~Pensky, ``Sparse
	convolutional neural networks,'' in \emph{Proceedings of the IEEE Conference
		on Computer Vision and Pattern Recognition (CVPR)}, 2015, pp. 806--814.
	
	\bibitem{srivastava-icml2015-movingmnist}
	N.~Srivastava, E.~Mansimov, and R.~Salakhutdinov, ``Unsupervised learning of
	video representations using lstms,'' in \emph{Proceedings of the
		International Conference on Machine Learning (ICML)}, 2015, pp. 843--852.
	
	\bibitem{zhang-aaai2017-trafficbj}
	J.~Zhang, Y.~Zheng, and D.~Qi, ``Deep spatio-temporal residual networks for
	citywide crowd flows prediction,'' in \emph{Proceedings of the AAAI
		conference on artificial intelligence(AAAI)}, 2017, pp. 1655--1661.
	
	\bibitem{ionescu--tpami2014-human3.6m}
	C.~Ionescu, D.~Papava, V.~Olaru, and C.~Sminchisescu, ``Human3.6m: Large scale
	datasets and predictive methods for 3d human sensing in natural
	environments,'' \emph{IEEE Transactions on Pattern Analysis and Machine
		Intelligence (TPAMI)}, vol.~36, no.~7, pp. 1325--1339, 2014.
	
	\bibitem{geiger-ijrr2013-kitti}
	A.~Geiger, P.~Lenz, C.~Stiller, and R.~Urtasun, ``Vision meets robotics: The
	{KITTI} dataset,'' \emph{International Journal of Robotics Research (IJRR)},
	vol.~32, no.~11, pp. 1231--1237, 2013.
	
	\bibitem{schuldt-icpr2004-kth}
	C.~Sch{\"{u}}ldt, I.~Laptev, and B.~Caputo, ``Recognizing human actions: {A}
	local {SVM} approach,'' in \emph{Proceedings of the Computational Vision and
		Active Perception Laboratory (CVAP)}, 2004, pp. 32--36.
	
	\bibitem{ranzato-arxiv2014-videomodeling}
	M.~Ranzato, A.~Szlam, J.~Bruna, M.~Mathieu, R.~Collobert, and S.~Chopra,
	``Video(language) modeling: A baseline for generative models of natural
	videos,'' \emph{arXiv preprint arXiv:1412.6604}, 2014.
	
	\bibitem{shi-nips2015-convlstm}
	X.~Shi, Z.~Chen, H.~Wang, D.~Yeung, W.~Wong, and W.~Woo, ``Convolutional {LSTM}
	network: {A} machine learning approach for precipitation nowcasting,'' in
	\emph{Advances in Neural Information Processing Systems (NeurIPS)}, 2015, pp.
	802--810.
	
	\bibitem{wang-tpami2023-predrnnv2}
	Y.~Wang, H.~Wu, J.~Zhang, Z.~Gao, J.~Wang, P.~S. Yu, and M.~Long, ``Predrnn:
	{A} recurrent neural network for spatiotemporal predictive learning,''
	\emph{IEEE Transactions on Pattern Analysis and Machine Intelligence
		(TPAMI)}, vol.~45, no.~2, pp. 2208--2225, 2023.
	
	\bibitem{oliu-eccv2018-frnn}
	M.~Oliu, J.~Selva, and S.~Escalera, ``Folded recurrent neural networks for
	future video prediction,'' in \emph{Proceedings of the European Conference on
		Computer Vision (ECCV)}, vol. 11218, 2018, pp. 745--761.
	
	\bibitem{su-nips2020-convttlstm}
	J.~Su, W.~Byeon, J.~Kossaifi, F.~Huang, J.~Kautz, and A.~Anandkumar,
	``Convolutional tensor-train {LSTM} for spatio-temporal learning,'' in
	\emph{Advances in Neural Information Processing Systems (NeurIPS)}, 2020.
	
	\bibitem{park-aaai2021-vidode}
	S.~Park, K.~Kim, J.~Lee, J.~Choo, J.~Lee, S.~Kim, and E.~Choi, ``Vid-ode:
	Continuous-time video generation with neural ordinary differential
	equation,'' in \emph{Proceedings of the AAAI conference on artificial
		intelligence (AAAI)}, 2021, pp. 2412--2422.
	
	\bibitem{kim-acmmm2021-dmee}
	N.~Kim and J.~Kang, ``Dynamic motion estimation and evolution video prediction
	network,'' \emph{IEEE Transactions on Multimedia (TMM)}, vol.~23, pp.
	3986--3998, 2021.
	
	\bibitem{lee-cvpr2021-lmc}
	S.~Lee, H.~G. Kim, D.~H. Choi, H.~Kim, and Y.~M. Ro, ``Video prediction
	recalling long-term motion context via memory alignment learning,'' in
	\emph{Proceedings of the IEEE Conference on Computer Vision and Pattern
		Recognition (CVPR)}, 2021, pp. 3054--3063.
	
	\bibitem{chen-cvpr2022-cpl}
	G.~Chen, W.~Zhang, H.~Lu, S.~Gao, Y.~Wang, M.~Long, and X.~Yang, ``Continual
	predictive learning from videos,'' in \emph{Proceedings of the IEEE
		Conference on Computer Vision and Pattern Recognition (CVPR)}, 2022, pp.
	10\,718--10\,727.
	
	\bibitem{yu-cvpr2022-mac}
	W.~Yu, W.~Chen, S.~Yin, S.~Easterbrook, and A.~Garg, ``Modular action concept
	grounding in semantic video prediction,'' in \emph{Proceedings of the IEEE
		Conference on Computer Vision and Pattern Recognition (CVPR)}, 2022, pp.
	3595--3604.
	
	\bibitem{devlin-naccl2019-bert}
	J.~Devlin, M.~Chang, K.~Lee, and K.~Toutanova, ``{BERT:} pre-training of deep
	bidirectional transformers for language understanding,'' in \emph{Proceedings
		of the 2019 Conference of the North American Chapter of the Association for
		Computational Linguistics: Human Language Technologies (NAACL-HLT)}, 2019,
	pp. 4171--4186.
	
	\bibitem{wei-cvpr2022-maskfit}
	C.~Wei, H.~Fan, S.~Xie, C.~Wu, A.~L. Yuille, and C.~Feichtenhofer, ``Masked
	feature prediction for self-supervised visual pre-training,'' in
	\emph{Proceedings of the IEEE Conference on Computer Vision and Pattern
		Recognition (CVPR)}, 2022, pp. 14\,648--14\,658.
	
	\bibitem{gao-nips2022-mcmae}
	P.~Gao, T.~Ma, H.~Li, Z.~Lin, J.~Dai, and Y.~Qiao, ``Mcmae: Masked convolution
	meets masked autoencoders,'' in \emph{Advances in Neural Information
		Processing Systems (NeurIPS)}, 2022.
	
	\bibitem{chen-ijcv2023-cae}
	X.~Chen, M.~Ding, X.~Wang, Y.~Xin, S.~Mo, Y.~Wang, S.~Han, P.~Luo, G.~Zeng, and
	J.~Wang, ``Context autoencoder for self-supervised representation learning,''
	\emph{International Journal of Computer Vision (IJCV)}, 2023.
	
	\bibitem{tong-nips2022-videomae}
	Z.~Tong, Y.~Song, J.~Wang, and L.~Wang, ``Videomae: Masked autoencoders are
	data-efficient learners for self-supervised video pre-training,'' in
	\emph{Advances in Neural Information Processing Systems (NeurIPS)}, 2022, pp.
	10\,078--10\,093.
	
	\bibitem{woo-cvpr2023-convnextv2}
	S.~Woo, S.~Debnath, R.~Hu, X.~Chen, Z.~Liu, I.~S. Kweon, and S.~Xie, ``Convnext
	{V2:} co-designing and scaling convnets with masked autoencoders,'' in
	\emph{Proceedings of the IEEE Conference on Computer Vision and Pattern
		Recognition (CVPR)}, 2023, pp. 16\,133--16\,142.
	
	\bibitem{liu-cvpr2022-convnext}
	Z.~Liu, H.~Mao, C.~Wu, C.~Feichtenhofer, T.~Darrell, and S.~Xie, ``A convnet
	for the 2020s,'' in \emph{Proceedings of the IEEE Conference on Computer
		Vision and Pattern Recognition (CVPR)}, 2022, pp. 11\,966--11\,976.
	
	\bibitem{tian-iclr2023-spark}
	K.~Tian, Y.~Jiang, Q.~Diao, C.~Lin, L.~Wang, and Z.~Yuan, ``Sparse and
	hierarchical masked modeling for convolutional representation learning,'' in
	\emph{Proceedings of the International Conference on Learning Representations
		(ICLR)}, 2023.
	
	\bibitem{howard-arxiv2017-mobilenet}
	A.~G. Howard, M.~Zhu, B.~Chen, D.~Kalenichenko, W.~Wang, T.~Weyand,
	M.~Andreetto, and H.~Adam, ``Mobilenets: Efficient convolutional neural
	networks for mobile vision applications,'' \emph{arXiv preprint
		arXiv:1704.04861}, 2017.
	
	\bibitem{dollar-cvpr2009-caltech}
	P.~Doll{\'{a}}r, C.~Wojek, B.~Schiele, and P.~Perona, ``Pedestrian detection:
	{A} benchmark,'' in \emph{Proceedings of the IEEE Conference on Computer
		Vision and Pattern Recognition (CVPR)}, 2009, pp. 304--311.
	
	\bibitem{villegas-iclr2017-mcnet}
	R.~Villegas, J.~Yang, S.~Hong, X.~Lin, and H.~Lee, ``Decomposing motion and
	content for natural video sequence prediction,'' in \emph{Proceedings of the
		International Conference on Learning Representations (ICLR)}, 2017.
	
	\bibitem{wang-tip2004-ssim}
	Z.~Wang, A.~C. Bovik, H.~R. Sheikh, and E.~P. Simoncelli, ``Image quality
	assessment: from error visibility to structural similarity,'' \emph{{IEEE}
		Transactions on Image Processing (TIP)}, vol.~13, no.~4, pp. 600--612, 2004.
	
	\bibitem{he-iccv2015-kaiming}
	K.~He, X.~Zhang, S.~Ren, and J.~Sun, ``Delving deep into rectifiers: surpassing
	human-level performance on imagenet classification,'' in \emph{Proceedings of
		the IEEE International Conference on Computer Vision (ICCV)}, 2015, pp.
	1026--1034.
	
	\bibitem{simth-ispp2019-onecycle}
	L.~N.Smith and N.~Topin, ``Super-convergence: Very fast training of neural
	networks using large learning rates,'' in \emph{Proceedings of the
		International Society for Optics and Photonics}, 2019.
	
	\bibitem{chang-tmm2022-stam}
	Z.~Chang, X.~Zhang, S.~Wang, S.~Ma, and W.~Gao, ``Stam: A spatiotemporal
	attention based memory for video prediction,'' \emph{IEEE Transactions on
		Multimedia (TMM)}, vol.~25, pp. 2354--2367, 2023.
	
	\bibitem{liu-iccv2017-dvf}
	Z.~Liu, R.~A. Yeh, X.~Tang, Y.~Liu, and A.~Agarwala, ``Video frame synthesis
	using deep voxel flow,'' in \emph{Proceedings of the IEEE International
		Conference on Computer Vision (ICCV)}, 2017, pp. 4473--4481.
	
	\bibitem{liang-iccv2017-dualmotiongan}
	X.~Liang, L.~Lee, W.~Dai, and E.~P. Xing, ``Dual motion {GAN} for future-flow
	embedded video prediction,'' in \emph{Proceedings of the IEEE International
		Conference on Computer Vision (ICCV)}, 2017, pp. 1762--1770.
	
	\bibitem{lotter-iclr2017-prednet}
	W.~Lotter, G.~Kreiman, and D.~D. Cox, ``Deep predictive coding networks for
	video prediction and unsupervised learning,'' in \emph{Proceedings of the
		International Conference on Learning Representations (ICLR)}, 2017.
	
	\bibitem{hao-cvpr2018-ctrlgen}
	Z.~Hao, X.~Huang, and S.~J. Belongie, ``Controllable video generation with
	sparse trajectories,'' in \emph{Proceedings of the IEEE Conference on
		Computer Vision and Pattern Recognition (CVPR)}, 2018, pp. 7854--7863.
	
	\bibitem{byeon-eccv2018-contextvp}
	W.~Byeon, Q.~Wang, R.~K. Srivastava, and P.~Koumoutsakos, ``Contextvp: Fully
	context-aware video prediction,'' in \emph{Proceedings of the European
		Conference on Computer Vision (ECCV)}, 2018, pp. 781--797.
	
	\bibitem{gao-iccv2019-dpg}
	H.~Gao, H.~Xu, Q.~Cai, R.~Wang, F.~Yu, and T.~Darrell, ``Disentangling
	propagation and generation for video prediction,'' in \emph{Proceedings of
		the IEEE International Conference on Computer Vision (ICCV)}, 2019, pp.
	9005--9014.
	
	\bibitem{jin-cvpr2020-stmfanet}
	B.~Jin, Y.~Hu, Q.~Tang, J.~Niu, Z.~Shi, Y.~Han, and X.~Li, ``Exploring
	spatial-temporal multi-frequency analysis for high-fidelity and
	temporal-consistency video prediction,'' in \emph{Proceedings of the IEEE
		Conference on Computer Vision and Pattern Recognition (CVPR)}, 2020, pp.
	4553--4562.
	
	\bibitem{geng-cvpr2022-vpcl}
	D.~Geng, M.~Hamilton, and A.~Owens, ``Comparing correspondences: Video
	prediction with correspondence-wise losses,'' in \emph{Proceedings of the
		IEEE Conference on Computer Vision and Pattern Recognition (CVPR)}, 2022, pp.
	3355--3366.
	
	\bibitem{jia-nips2016-dfn}
	X.~Jia, B.~D. Brabandere, T.~Tuytelaars, and L.~V. Gool, ``Dynamic filter
	networks,'' in \emph{Advances in Neural Information Processing Systems
		(NeurIPS)}, 2016, pp. 667--675.
	
	\bibitem{babaeizadeh-iclr2018-sv2p}
	M.~Babaeizadeh, C.~Finn, D.~Erhan, R.~H. Campbell, and S.~Levine, ``Stochastic
	variational video prediction,'' in \emph{Proceedings of the International
		Conference on Learning Representations (ICLR)}, 2018.
	
	\bibitem{lee-iclr2019-savp}
	A.~X. Lee, R.~Zhang, F.~Ebert, P.~Abbeel, C.~Finn, and S.~Levine, ``Stochastic
	adversarial video prediction,'' in \emph{Proceedings of the 7th International
		Conference on Learning Representations (ICLR)}, 2019.
	
	\bibitem{gao-iros2021-gridkeypoint}
	X.~Gao, Y.~Jin, Q.~Dou, C.~Fu, and P.~Heng, ``Accurate grid keypoint learning
	for efficient video prediction,'' in \emph{Proceedings of the International
		Conference on Intelligent Robots and Systems (IROS)}, 2021, pp. 5908--5915.
	
\end{thebibliography}

\ifCLASSOPTIONcaptionsoff
  \newpage
\fi

\end{document}